\documentclass[10pt,journal,compsoc]{IEEEtran}

\ifCLASSOPTIONcompsoc
\usepackage[nocompress]{cite}
\else
\usepackage{cite}
\fi

\ifCLASSINFOpdf
\else
\fi

\usepackage{amsmath}
\usepackage{bbm}
\usepackage{enumerate}
\usepackage{booktabs}
\usepackage{multirow}
\usepackage{graphicx}
\usepackage{subfigure}
\usepackage{color}
\usepackage{algorithm}
\usepackage{algorithmic}
\usepackage{hyperref}[colorlinks, linkcolor=black]

\begin{document}
	%
	\title{GANI: Global Attacks on Graph Neural \\ Networks via Imperceptible Node Injections}
	%
	%
	%
	%
	\author{Junyuan~Fang,~\IEEEmembership{Student Member,~IEEE}, Haixian Wen, Jiajing~Wu,~\IEEEmembership{Senior Member,~IEEE}, Qi Xuan,~\IEEEmembership{Senior Member,~IEEE}, Zibin Zheng,~\IEEEmembership{Senior Member,~IEEE}, and Chi K. Tse,~\IEEEmembership{Fellow,~IEEE}
		
		\IEEEcompsocitemizethanks{ 
			\IEEEcompsocthanksitem{Junyuan Fang and Chi K. Tse are with the Department of Electrical Engineering, City University of Hong Kong, Hong Kong SAR, China. (Email: junyufang2-c@my.cityu.edu.hk; chitse@cityu.edu.hk)}
			\IEEEcompsocthanksitem{Haixian Wen and Jiajing Wu are with the School of Computer Science and Engineering, Sun Yat-sen University, Guangzhou 510006, China. (Email: wenhx6@mail2.sysu.edu.cn; wujiajing@mail.sysu.edu.cn)}
			\IEEEcompsocthanksitem{Zibin Zheng is with the School of Software Engineering, Sun Yat-sen University, Zhuhai 519082, China. (Email: zhzibing@mail.sysu.edu.cn)}
			\IEEEcompsocthanksitem{Q. Xuan is   with the Institute of Cyberspace Security, College of Information Engineering, Zhejiang University of Technology, Hangzhou 310023, China. (E-mail: xuanqi@zjut.edu.cn)}
			\IEEEcompsocthanksitem{Corresponding Author: Jiajing Wu.}}
		\thanks{Digital Identifier: IEEE.xxx.xxx.xxxx.xxxxx.}
	}
	%
	%


	\IEEEtitleabstractindextext{%
		\begin{abstract}
			Graph neural networks (GNNs) have found successful applications in various graph-related tasks. However, recent studies have shown that many GNNs are vulnerable to adversarial attacks. In a vast majority of existing studies, adversarial attacks on GNNs are launched via direct modification of the original graph such as adding/removing links, which may not be applicable in practice. In this paper, we focus on a realistic attack operation via injecting fake nodes. The proposed \underline{G}lobal \underline{A}ttack strategy via \underline{N}ode \underline{I}njection (GANI) is designed under the comprehensive consideration of an unnoticeable perturbation setting from both structure and feature domains. Specifically, to make the node injections as imperceptible and effective as possible, we propose a sampling operation to determine the degree of the newly injected nodes, and then generate features and select neighbors for these injected nodes based on the statistical information of features and evolutionary perturbations obtained from a genetic algorithm, respectively. In particular, the proposed feature generation mechanism is suitable for both binary and continuous node features. Extensive experimental results on benchmark datasets against both general and defended GNNs show strong attack performance of GANI. Moreover, the imperceptibility analyses also demonstrate that GANI achieves a relatively unnoticeable injection on benchmark datasets.

		\end{abstract}
		
		\begin{IEEEkeywords}
			Graph Neural Networks, Graph Adversarial Attacks, Robustness, Node Injections, Unnoticeable Perturbations\end{IEEEkeywords}}

	\maketitle

	\IEEEdisplaynontitleabstractindextext

	%
	\IEEEpeerreviewmaketitle

	\IEEEraisesectionheading{\section{Introduction}\label{sec:introduction}}
	\IEEEPARstart{G}{raphs}, where nodes represent individuals and links represent relationships between individuals, are ubiquitous in real-world systems. With the remarkable success of deep graph learning, various powerful graph neural network (GNN) models have been proposed to deal with graph data-based tasks, such as node/graph classification, community detection, link prediction, and so on \cite{wu2020comprehensive, zhou2020graph, zhang2020deep, wu2020graph, jiang2021graph, 8294302, 9139346}.

	Recently, Goodfellow {\em et al.}\cite{DBLP:journals/corr/GoodfellowSS14} found that current neural networks are vulnerable to the unnoticeable adversarial samples generated by the attackers. Similarly, though GNNs have shown great potential on graph learning, researchers are also curious about the robustness of current GNNs when facing malicious attacks, namely, graph adversarial attacks \cite{sun2018adversarial,jin2020adversarial,chen2020survey,freitas2022graph}. 
	Prior work has proposed to evaluate the robustness of current GNNs against perturbations based on the gradient of classifiers \cite{zugner2018adversarial}, meta-learning \cite{zugner2018adversarial1}, reinforcement learning \cite{dai2018adversarial}, etc. Specifically, corresponding experiments have demonstrated the vulnerability of most current GNNs, even under small perturbations. 

	For the vast majority of existing studies on graph adversarial attacks \cite{zugner2018adversarial,dai2018adversarial,zugner2018adversarial1,chen2018fast,9245527}, a common assumption is that the attacker has the privilege to modify the original data, such as adding/removing links or modifying features of the original graph data. However, this assumption may not be valid in many realistic situations. Taking social networks as an example, it is rather difficult for an attacker to tamper with the relationship or the personal information of other users. Instead, an alternative strategy is to make an adversarial attacker generate some vicious nodes (i.e., adversarial nodes) and associate them with the original nodes to achieve the attacks. These kinds of adversarial attacks can be referred to as node injection attacks (NIA).

	In the NIA scenario, the attacker needs to solve two major problems. The first problem is to generate the feature of the adversarial nodes, while the other is to decide the neighbors of these newly injected nodes. Although a few NIA methods have been proposed, there still exist two main limitations. \textbf{(i) \em Feature Inconsistency}: Ignoring the diversity of feature spaces in different realistic networks, current work usually assumes the attacker generates binary features (0/1) to the new nodes. Other feature generation mechanisms like averaging features or combining with Gaussian distribution will also lead to inconsistent problems on ranges or types. \textbf{(ii) \em Budget Constraint}: Current NIA methods usually define the degree of each new node as a constant value such as the average degree of original nodes. Therefore, considering the above limitations, the NIAs in most existing work may cause obvious disturbance to the feature and topology domains of the original network, making the attacks easy to be detected.

	To address the above concerns and conduct effective and unnoticeable NIA, we propose a new NIA strategy GANI to achieve \underline{G}lobal \underline{A}ttacks via \underline{N}ode \underline{I}njections. We focus on global attacks (i.e., decreasing the global performance) of the node classification task via injecting some fake nodes into the original graph. Specifically, the proposed GANI can be divided into the following two parts. In the feature generation procedure, the statistical feature information of the original nodes is utilized to make our new nodes similar to the normal nodes. In the link generation procedure, we employ evolutionary algorithms to help find the optimal links which can obtain the best attack performance. Moreover, we utilize the decrease of node homophily \cite{pei2020geom,zhu2020beyond} to reduce the randomness of the optimal neighbor selections. Finally, we verify the superiority on both attack performance and imperceptibility of the proposed attack strategy in datasets with different feature types through extensive experiments. Our major contributions are as follows:

	\begin{itemize}
		
		\item We present new insights on promoting the imperceptibility of NIA from both topology and feature domains which were seldom discussed in previous studies.
		
		\item We propose a powerful NIA method by combining the statistical information-based feature generation step with evolutionary algorithm-based neighbor selection step on the global poisoning attack scenario.
		
		\item We conduct extensive experiments on several datasets with different types of features to verify the effectiveness and imperceptibility of the proposed method. 
	\end{itemize}
	
	The remainder of this work is organized as follows. Section \ref{sec:rw} discusses the related work of current attack and defense on graph neural networks. Some preliminaries about graph neural networks, node injection attacks, and the definition of homophily are given in Section \ref{sec:p}. Then, we introduce the setting of attack budgets, methods for feature generation and neighbor selection of injected nodes in Section \ref{sec:pm}. In Section \ref{sec:e}, we present the corresponding experiments and discussions on both attack performance and the imperceptible effect of perturbations in detail. Finally, we conclude our work in Section \ref{sec:c}.
	
	\section{Related Work}\label{sec:rw}
	In this work, we aim at achieving \textbf{global attacks}, which is to reduce the overall effectiveness of the models, rather than focus on influencing some target nodes. Moreover, we concentrate on \textbf{poisoning attacks} on graphs, for which we will apply our attack strategy on the retraining models, rather than on the fixed models with the same parameters before and after the attack (i.e., evasion attacks). Compared with evasion attacks, poisoning attacks will be more realistic but difficult due to the model retraining. In the following, we will briefly introduce current efforts on typical adversarial attacks and node injection attacks, respectively. Then we will briefly introduce current (defended) methods on graphs.

	\subsection{Typical Adversarial Attacks on GNNs} 
	
	In most existing studies, attacks are launched by modifying the links or features of the original nodes. Z$\ddot{u}$gner {\em et al.} \cite{zugner2018adversarial} proposed the first graph adversarial attack method, Nettack, to point out the security concern in current GNNs by utilizing the classification margin of the nodes predicted in different classes. Dai {\em et al.} \cite{dai2018adversarial} put forward a reinforcement learning-based strategy, RL-S2V, to achieve the targeted evasion attacks. In order to achieve global attacks, Z$\ddot{u}$gner and G{\"u}nnemann \cite{zugner2018adversarial1} further proposed Metattack (META for short in this work) by utilizing the meta gradient of loss functions. However, all of the above strategies are designed based on the assumption that the attackers can easily modify the original graphs, which is not realistic in the real life. Moreover, we also find that directly extending current methods to the node injection scenario can not achieve the same performance, which will be shown in our experiments later.
	
	\subsection{Node Injection Attacks on GNNs}
	
	To address the above concerns, some node injection attack methods have been proposed recently. Sun {\em et al.} \cite{sun2020adversarial} introduced a strong NIA strategy named NIPA which utilizes the deep Q-learning method to achieve the global poisoning attacks. Aiming at targeted poisoning attacks, Wang {\em et al.} \cite{wang2020scalable} proposed AFGSM by considering the gradient of loss function of targeted nodes. Zou {\em et al.} \cite{zou2021tdgia} introduced TDGIA, a global evasion injection method combining topological defective edge selection strategy and smooth feature generations. Moreover, Tao {\em et al.} \cite{tao2021single} proposed G-NIA by modeling the attack process via a parametric model to preserve the learned patterns and further be utilized in the inferring phase. Recently, Chen {\em et al.} \cite{chen2022understanding} proposed HAO by utilizing the homophily ratio of nodes as the unnoticeable constraint to further improve the unnoticeability of adversarial attacks.
	
	However, the two concerns (i.e., feature inconsistency and budget constraint) still have not been well addressed in prior work. Specifically, NIPA generates the adversarial features through the average features of all nodes plus Gaussian distribution. Since different datasets may have different types of features (i.e., binary or continuous features), and the features of each node usually will be an extremely sparse vector, the Gaussian noise may dominate the feature of new nodes, causing the situation that the features of newly injected nodes will be very different from those original nodes. Moreover, AFGSM simply generates binary features for the new nodes, while TDGIA only generates continuous features and fails to limit the feature-attack budget. Among them, only G-NIA solves the feature consistent problem. At the same time, none of the above work tackles the degree problems of the new nodes. Though HAO retains the homophily ratio of corresponding nodes during injections in a local perspective, it may break other global network properties or constraints such as degree distributions, the range/type of node features which can be easily detected.
	
	Therefore, current NIA methods may break the unnoticeable principle as the new nodes have a totally different range distribution or belong to different types (i.e., binary or continuous format) of features. They may also make the injection more obvious by assuming that all new nodes have the same degree. In order to make the NIA more realistic, we aim to address the above issues that exist in current strategies. Particularly, to make the adversarial attacks more general, we focus on a more practical and difficult scenario, namely global poisoning attacks.

	\begin{figure}[t]
		\subfigure[\textbf{Original Graph}]{
			\begin{minipage}[]{0.3\linewidth}
				\includegraphics[scale=0.25]{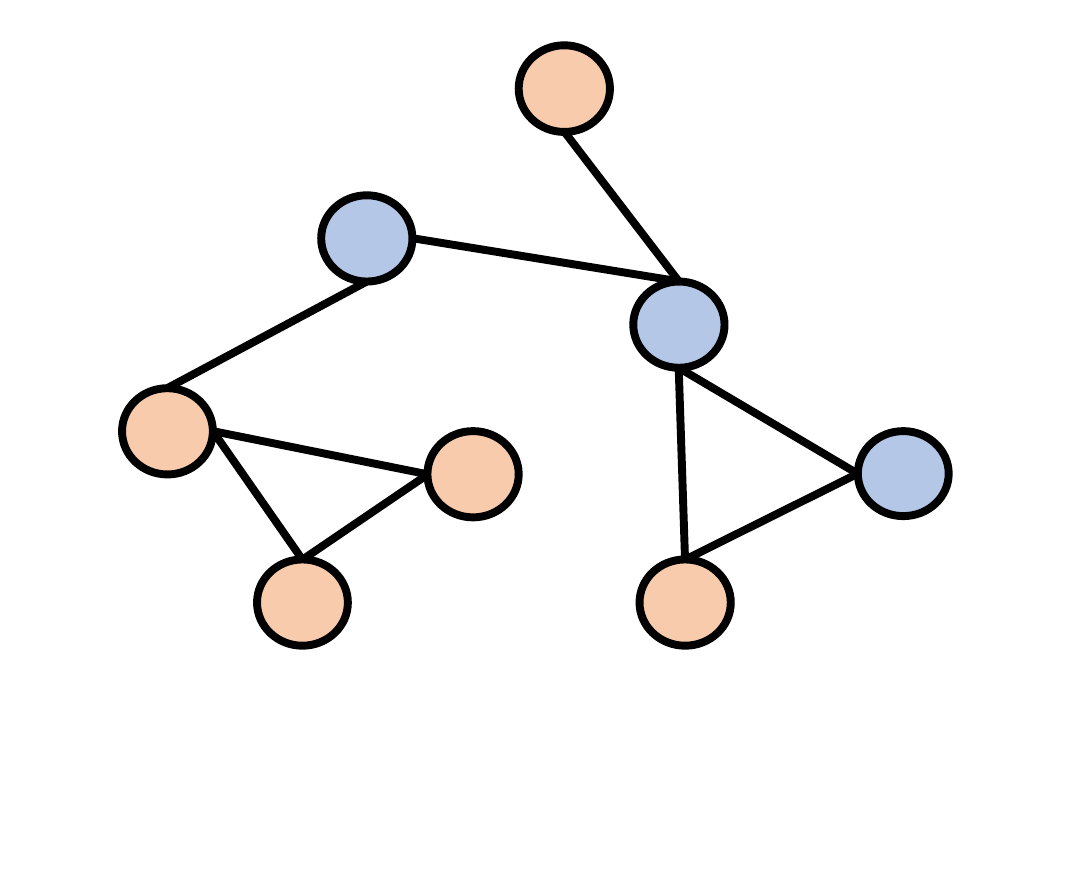}
			\end{minipage}%
			\label{fig:clean}
		}%
		\subfigure[\textbf{General Attacks}]{
			\begin{minipage}[]{0.3\linewidth}
				\includegraphics[scale=0.25]{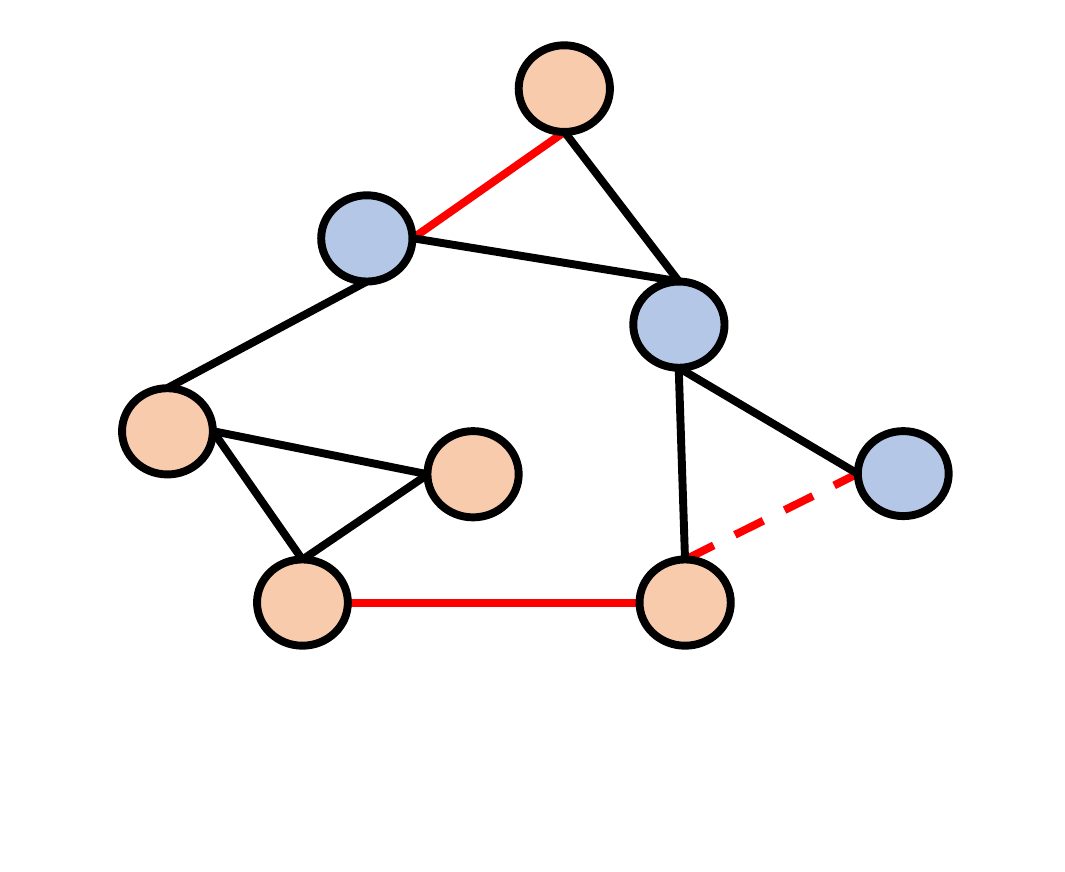}
			\end{minipage}%
		}%
		\subfigure[\textbf{Node Injections}]{
			\begin{minipage}[]{0.3\linewidth}
				\includegraphics[scale=0.25]{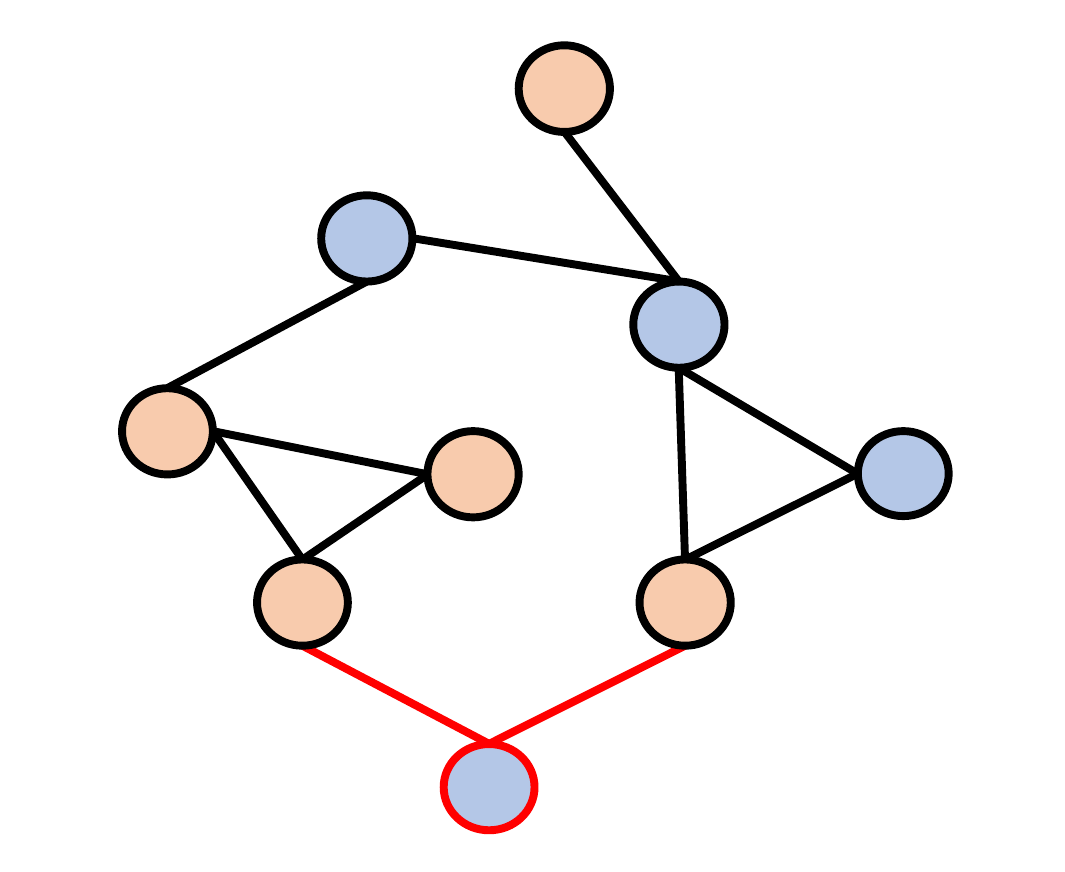}
			\end{minipage}%
		}%
		\centering
		\caption{Schematic diagram of general attacks and node injection attacks. Colors represent labels; red solid and dashed lines indicate adding and removing links, respectively.}
		\label{fig:schematic}
	\end{figure}

	\subsection{Defense on GNNs} 
	
	Standard GNNs usually utilize the neighborhood aggregation mechanism to help obtain a better representation of the central node from its neighbors, such as classic graph convolution network (GCN) \cite{kipf2017semi}, graph attention network (GAT) \cite{velivckovic2018graph} by giving different aggregation weights to different neighbors via a masked self-attention mechanism, simplifying graph convolution network (SGC) \cite{wu2019simplifying} by removing the activation functions in the middle layers, etc. 
	
	Since current GNNs are vulnerable to adversarial attacks, some other models are proposed to improve the robustness of GNNs through the pre-processing of data, re-designing the aggregation functions, etc. For example, Wu {\em et al.} \cite{ijcai2019-669} came up with Jaccard by removing the links with relatively lower feature similarity before employing the GNNs. Zhu {\em et al.} \cite{zhu2019robust} designed RGCN via modeling the low-dimension representation of nodes as Gaussian distribution and further employing a variance-based attention mechanism as the aggregation function to reduce the influence of adversarial samples. Jin {\em et al.} \cite{jin2021node} introduced SimPGCN by adaptively preserving both the feature similarity and structural similarity during the aggregation process. Moreover, Chen {\em et al.} \cite{ijcai2021p310} proposed novel aggregators, {\em TMean} and {\em Median}, based on the breakdown point to discard the possible outlier features during the neighborhood aggregation on evasion attack scenario.  In our experiments, we also further verify the attack performance of the proposed method in these defended GNNs.

	\begin{table}[]
		\centering
		\caption{Frequently used notations in this paper and their corresponding definitions or descriptions.}
		\resizebox{\linewidth}{!}{
			\begin{tabular}{c|l}
				\bottomrule
				\textbf{Notations} & \textbf{Definitions or Descriptions} \\ \hline\hline
				$G$    & Graph data \\ 
				$A$    & Adjacency matrix of the graph\\
				$X$    & Feature matrix of the graph\\
				$f_{\theta}$    & The graph neural network model with parameter $\theta$\\
				$W$    & Trainable weight matrix\\
				$Z$    & Prediction probability of all nodes. \\
				$G^{\prime}$    & Perturbed graph data\\
				$A^{\prime}$    & Perturbed adjacency matrix of the graph\\
				$X^{\prime}$    & Perturbed feature matrix of the graph\\
				$A_{P}$    & Adjacency matrix between fake nodes and normal nodes\\
				NH    & Node homophily\\
				$\Delta F$    & Feature-attack budget\\
				$\Delta L$    & Link-attack budget list for all injected nodes\\
				$n_{in}$    & Number of injected nodes\\
				$\alpha$    & Candidate rate\\
				$p_c$    & Crossover rate\\
				$p_m$    & Mutation rate\\
				DNH    & Decrease of node homophily\\
				TDNH    & Total decrease of node homophily in an individual\\
				$P$    &  Population of genetic algorithm\\
				$I$    & Individual in population\\
				$L^{\prime}$    & Randomly given label for the injected node\\	
				\bottomrule
		\end{tabular}}
		\label{table:notations}
	\end{table}

	\section{Preliminaries}\label{sec:p}
	In this section, we will give some preliminaries on graph neural networks and node injection attacks, together with the definition of node homophily. See Table \ref{table:notations} for frequently used notations.
	
	\subsection{Graph Neural Networks}
	Generally, we use $G=(A, X)$ to denote the graph, where $A$ and $X$ are the adjacency matrix and feature matrix, respectively. Assuming $G$ has $n$ nodes and each of them has $d$ features, the adjacency matrix $A \in \{0,1\}^{n \times n}$ represents the connection relationship between each node-pair, where $A_{ij} = 1$ means there is a link between node $i$ and $j$, and 0 otherwise. The feature matrix $\{X\}^{n \times d}$ denotes the features of each node, and $X_i$ represents the feature vector of node $i$. As mentioned before, the feature vector could be either binary or continuous type. Following the classic work on node classification \cite{kipf2017semi}, a general two-layer graph convolution network can be represented by
	\begin{equation}\label{z_gcn}
		Z = f_{\theta}(A,X)=\text{softmax}(\hat{A} \sigma{(\hat{A}XW^{(1)})} W^{(2)}),
	\end{equation}
	where $\hat{A} = \tilde{D}^{-\frac{1}{2}}(A+I)\tilde{D}^{-\frac{1}{2}}$, $\tilde{D}$ and $I$ represent the diagonal degree matrix and identity matrix, respectively. $\sigma{(\cdot)}$ is the activation function like ReLU, and $W$ is the weight matrix to be optimized. Specifically, $Z_{v, c}$ denotes the probability that node $v$ belongs to class $c$. The optimization goal of node classification is to assign the node a higher probability to the true class, i.e.,
	
	\begin{equation}
		\min_{\theta} L = - \sum_{v \in V_{\rm train}} \ln{Z_{v,c_v}},
	\end{equation}
	where $c_v$ indicates the ground truth label of node $v$, $V_{\rm train}$ represents the nodes of training set, and $\theta = \{W^{(1)}, W^{(2)}\}$ is the learning parameter to be optimized.

	\subsection{Node Injection Attacks}
	Instead of directly modifying the original adjacency matrix $A$ and feature matrix $X$, NIA will inject some malicious nodes into the original graph and generate the features of these injected nodes, and then connect them to the original graph to conduct the attack. The adversarial graph after being injected $n_{in}$ fake nodes is given as
	
	\begin{equation}\label{matrix_a}
		A^{'} = 	\left[ \begin{array}{cc}
			A & A_{p}^{T}\\
			A_{p} & O
		\end{array}
		\right ],
	\end{equation}

	\begin{equation}\label{matrix_x}
		X^{'} = 	\left[ \begin{array}{c}
			X\\
			X_{p}
		\end{array}
		\right ],
	\end{equation}
	where matrices $\{A_p\}^{n_{in} \times n}$  and $\{A_p^{T}\}^{n \times n_{in}}$ represent the adversarial matrices between the fake nodes and normal nodes and corresponding transpose matrix, respectively. $\{X_p\}^{n_{in} \times d}$ is the new generated features of the new nodes.

	Specifically, we focus on global poisoning attacks on node classification, and the final goal of the attacker is to decrease the overall classification accuracy of the targeted GNNs, which can be represented as 
	
	\begin{equation}\label{attack_loss}
		\begin{split}
			&\max\sum_{v \in V_{\rm test}}    \mathbbm{1}{(\arg \max{\ln{{f_{\theta^{*}}(A^{'},X^{'})}_{v}}} \not = c_v)},
			\\
			&\:\mathrm{s.t.} \theta^{*} =\arg \max \limits_{\theta} \sum_{v \in V_{\rm train}} \ln{{f_{\theta}(A^{'},X^{'})}_{v,c_v}}
		\end{split}
	\end{equation}
	where $\theta^{*}$ indicates the optimal training parameters trained on the perturbed graph, $\arg \max{\ln{{f_{\theta^{*}}(A^{'},X^{'})}_{v}}}$ is the predicted label of node $v$ based on $\theta^{*}$, and $\mathbbm{1}{(\cdot)}$ is an indicator function with output 1 when the argument is true, and 0 otherwise. 

	\subsection{Node Homophily}
	Similar to some previous studies \cite{pei2020geom,zhu2020beyond}, we define \textbf{node homophily (NH)} of a specific node based on the ratio of neighbors whose label is the same as the central nodes. The detailed calculation of NH and the average NH of the whole graph is given by
\begin{eqnarray}
		{\rm NH} &=& \frac{\# \ {\rm neighbors \ with \ same \ label}}{\# \ {\rm all \ neighbors}}, \\
		{\rm Average \ NH} &=& \frac{1}{n} \sum_{i=1}^{n} {\rm NH}_i,
\end{eqnarray}
where $n$ is the total number of nodes in the graph.
	
	To get a better understanding of node homophily, we analyze average NH of the largest connected component of four benchmark datasets, whose statistical information is given in Table \ref{table: dataset} in Section 5. As shown in Table \ref{table:homophily}, we can observe that all these datasets have an extremely high average NH closing to 0.9, meaning that most neighbors will have the same label as the central nodes. The above finding is also consistent with the design of the neighborhood aggregation mechanism on GNNs which utilizes the information of neighbors to help the central nodes obtain a better representation. Therefore, we try to utilize node homophily to design our attack strategy in the next section.

	\begin{table}[]
		\centering
		\caption{Average node homophily of the largest connected component of four datasets.}
		
		\begin{tabular}{ccccc}
			\bottomrule
			\textbf{Datesets} & \textbf{Cora} & \textbf{Citeseer} & \textbf{Cora-ML} & \textbf{Pubmed} \\ \hline\hline
			Average NH    & 0.8787        & 0.9136            & 0.8720           & 0.8948          \\ \bottomrule
		\end{tabular}
		\label{table:homophily}
	\end{table}

	\section{Proposed Method}\label{sec:pm}
	
	In this section, we describe the proposed method for solving the following issues. First, to keep our attacks from being detected, we need to limit our adversarial perturbations on both structure and feature domains to reasonable budgets. Then, we need to solve the two major challenges in NIA, including feature generation and neighbor selection.

	\subsection{Attack Budgets}
	In order to make the attacks as unnoticeable as possible, we need to define the attack limitation appropriately. Specifically, in the NIA scenario, we decide both the feature-attack budget $\Delta F$ and link-attack budget $\Delta L = \{\Delta L_0, \Delta L_1, \cdots , \Delta L_{n_{in}-1}\}$ as we will inject some fake nodes into the original graph. For the feature-attack budgets, since the features of nodes are usually extremely sparse vectors, we follow the setting of the previous work \cite{wang2020scalable} by assigning the average number of non-zero features of normal nodes to the newly injected nodes.

	For the link-attack budget, previous work usually assigns the average degree of original nodes to all new nodes. Yet we believe such settings will cause an obvious change to the graph's structural properties as many new nodes having the same degree will be injected at the same time, which can be easily detected. Therefore, to keep a higher similarity in the structural properties between the graph before and after the attacks, we propose a degree sampling operation to decide the link-attack budget of each newly injected node. Specifically, we randomly select an original node firstly, then assign the degree of this node to the current new node, and repeat this procedure until all $n_{in}$ nodes have been assigned. Particularly, we set the maximal link-attack budget of each injected node to $2 \times average \ degree$, aiming to avoid the extremely large perturbations on a single node. Since the degree with a higher frequency in the original graph will have a higher probability to be sampled in the above procedure, the similarity of graphs before and after the attacks can be retained to a great extent.

	\subsection{Feature Generation of Injected Nodes}
	To avoid the attacks being detected, we prefer the newly injected nodes to be similar to the original nodes. Therefore, from the perspective of features, we employ a statistical operation to generate the adversarial features. During the feature generation of each new node, we first assign a targeted label $L^{'}$ randomly, and then calculate the total non-zero appearance of features in each index among $d$ dimensions of all original nodes whose label belongs to class $L^{'}$. Further, we choose the top $\Delta F$ feature indices with the highest appearance as the non-zero indices of the current injected node. After that, for a specific feature index ${\Delta F}_i$ of new node $k$, the specific feature value will be calculated as
	\begin{equation}\label{adv_f}
		{X}^{'}[k,{\Delta F}_i] = \frac{\sum_{Y[v] = L^{'}} X[v,{\Delta F}_i]}{\sum_{Y[v] = L^{'}}{ \mathbbm{1}{(X[v,{\Delta F}_i] \not = 0)} } }, k \in [n,n+n_{in}),
	\end{equation}
	where $Y$ is the label matrix and $X$ is the original feature matrix. 
	
	It is worth noting that the above feature generation mechanism is suitable for both binary and continuous features. For binary features, the calculated feature values on the corresponding non-zero indices $X_{k,{\Delta F}_i}^{'}$ will also be 1. Since $X$ is usually a sparse matrix,  (\ref{adv_f}) can promise the newly injected features having the same range of value as normal nodes because averaging is only done by the non-zero features in this index, rather than all nodes. After the above feature generation operation, the newly generated node will have a relatively higher similarity with the normal nodes with the same targeted label in the feature domain. In addition, we also can apply some randomness to this feature generation procedure to make it more unnoticeable.


	\subsection{Neighbor Selection of Injected Nodes}
	After generating the adversarial features to the newly injected nodes, our next challenge is to assign appropriate neighbors to achieve the best attack performance. According to  (\ref{attack_loss}), the principal goal of the proposed model is to decrease the overall prediction accuracy of testing nodes. In other words, we want to figure out what kind of changes in the adversarial adjacency matrix $A^{'}$ and feature matrix $X^{'}$ can achieve the best attack performance. Based on the analysis of Table \ref{table:homophily}, we know that higher node homophily usually helps the GNNs obtain a better performance in the downstream task as the neighbors have a positive effect on obtaining the representation of central nodes during aggregations. Therefore, an intuitive idea is to decrease the homophily of nodes, thereby influencing the neighborhood aggregation procedure of GNNs.


	For a specific injected node, we need to select its neighbors that can improve the optimization of (\ref{attack_loss}). Interestingly, the neighbor selection can be modeled as a combinatorial optimization problem. Evolutionary algorithms designed by imitating the biological evolution processes have proved to be an effective strong tool to solve complicated combinatorial optimization problems \cite{bliss2014evolutionary, chen2018optimal, fang2020revealing, fang2021enhancing}. In the graph adversarial learning domain, previous work also utilizes the evolutionary algorithms to achieve optimal perturbations on the graph modification attacks \cite{dai2018adversarial}, community deceptions \cite{chen2019ga,9246590}, link hiding in social networks \cite{yu2019target}, etc.

	Therefore, we employ the classic genetic algorithm (GA) \cite{holland1992genetic} to help find the near-optimal neighbors for each newly injected node. Similar to the previous work \cite{zugner2018adversarial,li2020adversarial}, we also train a surrogate model $M_{s}$ based on a two-layer SGC model by simplifying  (\ref{z_gcn}) through removing the activation function in the first layer, i.e.,
	\begin{algorithm}[tb]
		\caption{GANI}
		\label{alg}
		\textbf{Input}: Original graph $G=(A,X)$, injected node numbers $n_{in}$, candidate rate $\alpha$, crossover rate $p_c$, mutation rate $p_m$\\
		\textbf{Output}: Final adversarial graph $G^{'} = (A^{'}, X^{'})$
		
		\begin{algorithmic}[1] 
			\STATE Train a surrogate model $M_s$ based on the original $G$.
			\STATE $\Delta L \gets $ Generate the link-attack budget of $n_{in}$ based on degree sampling operation.
			\FOR {each i $\in [{0, n_{in} - 1}]$}
			\STATE $L^{'} \gets$ Randomly assign a targeted label to current node $v_i$.
			\STATE ${X_p}_{i} \gets$ Calculate the adversarial feature vector for $v_i$ according to  (\ref{adv_f}).
			\STATE $t \gets 0$
			\STATE $candidates \gets Candidates\_selection (L^{'}, \alpha)$
			\STATE $P^{t} \gets Population\_initialization (candidates, \Delta L_i) $
			\WHILE {$t < $ maximal iterations}
			\STATE $P^{t} \gets Crossover (P^{t}, P_c)$
			\STATE $P^{t} \gets Mutation (P^{t}, P_m)$
			\STATE $F^{t}_{fit} \gets Fitness\_calculation (P^{t}, M_s)$
			\STATE $P^{t+1} \gets Elite\_selection (P^{t}, F^{t}_{fit})$
			\STATE $t \gets t+1$
			\ENDWHILE
			\STATE ${A_p}_{i} \gets$ Transform the optimal link-attack solution among $P^{t}$ to the matrix format.
			\STATE Update $A^{'}, X^{'}$ from ${A_p}_{i}, {X_p}_{i}$ according to (\ref{matrix_a}) and (\ref{matrix_x}).
			\ENDFOR
			\RETURN $G' = (A^{'}, X^{'})$
		\end{algorithmic}
	\end{algorithm}


	\begin{equation}
		Z^{s} = \text{softmax}(\hat{A}\hat{A}XW^{(1)}W^{(2)}) = \text{softmax}(\hat{A}^2XW^{'}),
	\end{equation}
	where $W^{'} = W^{(1)}W^{(2)}$.
	
	For the fitness function of GA, we utilize the number of misclassified nodes in the pre-trained surrogate model $M_s$ as an indicator, i.e.,
	
	\begin{equation}\label{f_fit}
		f_{\rm fit} = \sum_{v \in V_{\rm test}} \mathbbm{1}{(\arg \max{\ln{{f^{'}(A^{'},X^{'})}_{v}}} \not = c_v)},
	\end{equation}
	where $f^{'}(A^{'},X^{'})$ refers to the latest probability output by applying corresponding $A^{'}$ and $X^{'}$ under the surrogate model $M_s$. Also, $c_v$ could be the ground truth or the predicted label from surrogate models under different settings.

	\begin{figure*}[t]
		\centering
		\includegraphics[width=1\textwidth]{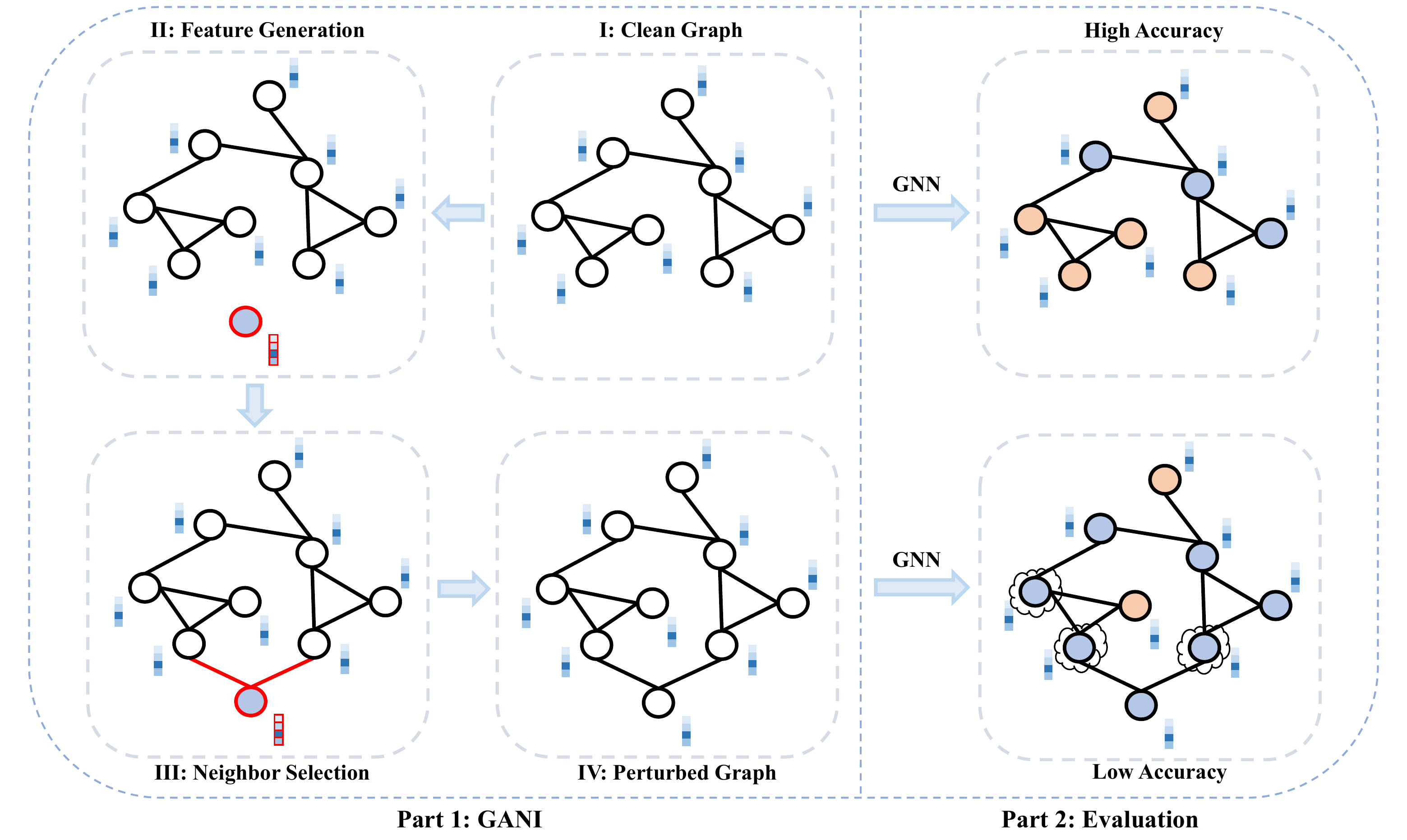} 
		\caption{A systematic framework of GANI and corresponding evaluations. The red marks indicate the corresponding generated fake node including both features and neighbors. The colors of nodes represent the classes, and the cloud-shaped circle means a wrong classification of the node.}
		\label{fig:framework}
	\end{figure*}

	In the experiments, we find that the outputs of different $A^{'}$ in  (\ref{f_fit}) can only differ slightly. In other words, there may be numerous different $A^{'}$'s that have the same value of $f_{\rm fit}$. If we randomly select one of them as the optimal adversarial adjacency, our method may be unstable due to the randomness on neighbor selection. Therefore, we introduce another sorting index, the decrease of node homophily (DNH), to reduce the randomness of the above selection. The definition of DNH of a specific node $i$ is as 
	\begin{equation}\label{f_dnh}
		{\rm DNH}_i = {\rm NH}^{\rm before}_i - {\rm NH}^{\rm after}_i,
	\end{equation}
	where ${\rm NH}^{\rm before}_i$ and ${\rm NH}^{\rm after}_i$ refer to the node homophily before and after node $i$ being attacked, respectively. Moreover, for a specific individual $I = \{(n_v, n_a), (n_v, n_b), \cdots, (n_v, n_k)\}$ that consist of $k$ injected links in GA, the total DNH (labeled as TDNH) of individual $I$ is given as
	\begin{equation}\label{f_tdnh}
		{\rm TDNH}_I = {\rm DNH}_{n_a} + {\rm DNH}_{n_b} + \cdots + {\rm DNH}_{n_k},
	\end{equation}
	where ${\rm DNH}_{n_{(\cdot)}}$ is the specific decrease of node homophily of node $n_{(\cdot)}$.
	

	Following the definition of TDNH, we further define a two-level sorting mechanism by combining the original fitness function $f_{\rm fit}$ and the total decrease of node homophily TDNH in each individual. For instance, for two different $A^{'}$ with the specific value (${f_{\rm fit}}_1$, ${\rm TDNH}_1$) and (${f_{\rm fit}}_2$, ${\rm TDNH}_2$), the proposed sorting mechanism will first order them based on the priority of original fitness value if ${f_{\rm fit}}_1 \not= {f_{\rm fit}}_2$. However, if ${f_{\rm fit}}_1 = {f_{\rm fit}}_2$, we will retain the individual with a higher TDNH. Based on this principle, the major steps of GA for a specific node $n_v$ are as follows.
	

	\begin{enumerate}
		\item \textbf{Parameter Definition.} We define a specific individual as $I = \{(n_v, n_a), (n_v, n_b), \cdots , (n_v, n_k)\}$ where $(n_v, n_{(\cdot)})$ is a link connecting the adversarial node $n_v$ with a normal node $n_{(\cdot)}$. The population is $P = \{I_0, I_1, I_2,\cdots\}$, candidate rate is $\alpha$, crossover rate and mutation rate are $p_c$ and $p_m$, respectively.   
		\item \textbf{Candidates Selection.} For current node $n_v$ with the targeted label $L^{'}$, as the searching space is extremely large due to the inclusion of all possible nodes as $n_{(\cdot)}$, we first remove the nodes having label $L^{'}$ as we want to decrease the node homophily. Then we employ single-link attacks under the guide of $f_{\rm fit}$ and TDNH. We will only retain the nodes in the top $\alpha$ as the final candidates which may be selected later.
		\item \textbf{Population Initialization.} From the obtained candidates, we generate an initial population by randomly selecting the node from candidates as node $n_{(\cdot)}$ in each individual.
		\item \textbf{Crossover and Mutation.} Under the crossover rate $p_c$ and mutation rate $p_m$, we will exchange a part of adversarial links between every two individuals in the crossover operation, and replace an adversarial link with a new randomly generated link in each individual in the mutation operation.
		
		\item \textbf{Fitness Calculation.} We inject each individual to the original graph by constructing current adversarial adjacency matrix $A^{'}$. Together with the features $X^{'}$ generated from feature generation step, we can obtain the fitness list $F_{\rm fit}$ by recording each $f_{\rm fit}$ and TDNH calculated through (\ref{f_fit}) and (\ref{f_tdnh}).
		
		\item \textbf{Elite Selection.} Finally, to maintain the scale of $P$, we utilize the tournament selection mechanism to retain the better individuals. The individuals with higher $f_{\rm fit}$ and TDNH will be more likely to be retained.
	\end{enumerate}
	Specifically, Steps 4 to 6 will repeat until the predefined maximum iteration is exceeded.

	\subsection{Overall Framework}
	To sum up, the overall framework of the proposed GANI on global poisoning attacks on GNNs consists of the GANI attack phase and evaluation phase, as illustrated in Fig. \ref{fig:framework}. During the attack phase, we first assign the link-attack budgets to each newly injected node through degree sampling operation, while the feature-attack budget is restricted by the average non-zero feature indices of original nodes. Then, we generate the specific features for the current adversarial node based on the proposed statistical information of features in the feature generation step. Next, we will utilize GA to find the near-optimal neighbors on neighbor selection step. This procedure will repeat until all $n_{in}$ adversarial nodes have successfully been injected, so the proposed GANI is a sequential generation method rather than a one-time injection, which will be more flexible if we want to increase the number of injection nodes. Finally, in the evaluation phase, we input the constructed adversarial graph to the GNNs for training as we focus on poisoning attacks, and further verify the attack performance of GANI.

	\section{Experiments}\label{sec:e}
	In this section, we verify the effectiveness of GANI via comprehensive experiments. we will first introduce the corresponding setups including the statistics of datasets, detailed baseline methods, the GNNs that will be attacked, and parameter settings. Then, we present corresponding experimental results and analysis to verify the performance of the proposed method.
	
	\subsection{Experimental Setup}
	\subsubsection{Datasets} 
	We choose four representative benchmark datasets including Cora, Citeseer, Pubmed \cite{sen2008collective}, and Cora-ML \cite{mccallum2000automating} to conduct the experiments. Among them, Cora and Citeseer have binary features, while the other two have continuous features. Adopting the setting of previous work \cite{zugner2018adversarial}, we use the largest connected component of these datasets in our experiments. The statistics are shown in Table \ref{table: dataset}.

	\begin{table}[]
		\centering
		\renewcommand\arraystretch{1.1}
		\caption{Statistics of datasets. Last column indicates whether the corresponding dataset has binary features.}
		
		\begin{tabular}{c|rrrc}
			\bottomrule
			\textbf{Datesets} & \textbf{\# Nodes} & \textbf{\# Links} & \textbf{\# Features} & \textbf{Binary} \\ \hline\hline
			Cora              & 2,485           & 5,069           & 1,433              & Y                        \\
			Citeseer          & 2,100           & 3,668           & 3,703              & Y                        \\ 
			Cora-ML           & 2,810           & 7,981           & 2,879              & N                        \\ 
			Pubmed            & 19,717          & 44,324          & 500               & N                        \\ \bottomrule
		\end{tabular}
		\label{table: dataset}
	\end{table}

	\begin{table*}[]
		\centering
		\renewcommand\arraystretch{1.2}

		\caption{Accuracy of the four GNNs with injecting 5\% nodes. GANI(L) refers to the results of GANI with ground truth label. The best result of each row (except for method GANI(L)) is boldfaced. Here `OOM' denotes out of memory.}
		\setlength{\tabcolsep}{3mm}{
		\begin{tabular}{c c c c c c c c c c c}
			\bottomrule
			\textbf{Datesets}         & \textbf{Attack} & \multicolumn{1}{l}{\textbf{Clean}} & \textbf{Nettack} & \textbf{META} & \textbf{AFGSM} & \textbf{G-NIA} & \textbf{TDGIA} & \textbf{GANI} & \textbf{GANI(L)}\\ \bottomrule\bottomrule

			\multirow{4}{*}{Cora} & GCN & 0.8360 & 0.8183 & 0.8100 & 0.8240 & 0.8132 & 0.8266 & \textbf{0.7725} & 0.7492\\ \cline{2-10} 
			& SGC & 0.8340 & 0.8121 & 0.8129 & 0.8161 & 0.8010 & 0.8255 & \textbf{0.7806} & 0.7539 \\ \cline{2-10} 
			& Jaccard & 0.8355 & 0.8184 & 0.8158 & 0.8190 & 0.8152 & 0.8269 & \textbf{0.7835} & 0.7546 \\ \cline{2-10} 
			& SimPGCN & 0.8294 & 0.8118 & 0.8094 & 0.8116 & 0.8114 & 0.8155 & \textbf{0.7884} & 0.7792 \\ 
			\bottomrule

			\multirow{4}{*}{Citeseer} & GCN & 0.7287 & 0.7080 & 0.7149 & 0.7173 & 0.7146 & 0.7174 & \textbf{0.7027} & 0.6870 \\ \cline{2-10} 
			& SGC & 0.7248 & 0.7075 & 0.7125 & 0.7079 & \textbf{0.7008} & 0.7068 & 0.7051 & 0.6862 \\ \cline{2-10} 
			
			& Jaccard & 0.7294 & 0.7096 & 0.7201 & 0.7176 & 0.7150 & 0.7108 & \textbf{0.7053} & 0.6846 \\ \cline{2-10} 
			& SimPGCN & 0.7547 & 0.7387 & 0.7273 & 0.7415 & 0.7438 & 0.7421 & \textbf{0.7080} & 0.7022 \\ 
			\bottomrule

			\multirow{4}{*}{Cora-ML} & GCN & 0.8536 & 0.8423 & 0.8529 & 0.8519 & 0.8439 & 0.8422 & \textbf{0.8082} & 0.7830 \\ \cline{2-10} 
			& SGC & 0.8480 & 0.8316 & 0.8450 & 0.8430 & 0.8375 & 0.8359 & \textbf{0.8076} & 0.7744 \\ \cline{2-10} 
			
			& Jaccard & 0.8539 & 0.8420 & 0.8523 & 0.8515 & 0.8462 & 0.8427 & \textbf{0.8094} & 0.7832 \\ \cline{2-10} 
			& SimPGCN & 0.8485 & 0.8344 & 0.8432 & 0.8455 & 0.8463 & 0.8393 & \textbf{0.8103} & 0.7940 \\ 
			\bottomrule
			
			\multirow{4}{*}{Pubmed} & GCN & 0.8649 & 0.8498 & OOM & 0.8568 & 0.8710 & 0.8568 & \textbf{0.8392} & 0.8098 \\ \cline{2-10} 
			& SGC & 0.8516 & 0.8315 & OOM & 0.8422 & 0.8501 & 0.8423 & \textbf{0.8119} & 0.7700 \\ \cline{2-10} 
			
			& Jaccard & 0.8652 & 0.8502 & OOM & 0.8581 & 0.8721 & 0.8566 & \textbf{0.8406} & 0.8130 \\ \cline{2-10} 
			& SimPGCN & 0.8788 & \textbf{0.8730} & OOM & 0.8751 & 0.8764 & 0.8746 & \textbf{0.8730} & 0.8671 \\ 
			\bottomrule\bottomrule
			
			\multicolumn{2}{c}{Average Ranking} & / & 2.69 & 6.25 & 4.50 & 3.88 & 4.06 & \textbf{1.06} & / \\
			\bottomrule
		\end{tabular}}
		
		\label{table:gnns}
	\end{table*}

	\subsubsection{Baselines}
	For the baselines, we mainly compare our GANI with the extension of general targeted and global attack methods on NIA scenarios, and recent targeted and global NIA methods. Specifically, the compared methods can be divided into the following three groups. The first group is the extension of two strong general attack methods, including \textbf{Nettack} \cite{zugner2018adversarial} and \textbf{META} \cite{zugner2018adversarial1}. The second group is the targeted NIA methods including \textbf{AFGSM} \cite{wang2020scalable} and \textbf{G-NIA} \cite{tao2021single}, and the last group is the global NIA method, \textbf{TDGIA} \cite{zou2021tdgia}. We do not compare our method with \text{NIPA} \cite{sun2020adversarial} as no official codes are available.

	Since current attack methods are not directly designed to the global poisoning attacks under NIA scenarios, we need to make some necessary modifications to extend their default designs, for example, from general attacks to node injection attacks, from the targeted attacks to global attacks, etc. To ensure fairness, for the feature generation of newly injected nodes, we will randomly copy the features of the original nodes to the newly generated nodes if the corresponding baselines do not solve the feature consistent problem under a given feature-attack budget.

	The detailed description of baselines and corresponding extended designs of them are as follows. 
	
	\begin{enumerate}
		
		\item{\textbf{Nettack.}} Nettack is a powerful targeted attack method based on modifying the original graph and the feedback of classification margins. To extend the original Nettack to the NIA scenario, we first randomly sample $n_{in}$ feature vectors and corresponding labels of original nodes as the newly injected nodes, then initialize the original neighbors following the generation of the BA scale-free model \cite{barabasi2003scale} by giving a higher connected probability to the original nodes with a higher degree. Moreover, as Nettack is a targeted attack method while we focus on global attacks, we borrow the idea from the comparison between Nettack to the global attack method META (which will be introduced later). Specifically, during the injection of each node, we randomly choose a node in the test set as the targeted node, then employ the original Nettack but limit the adversarial link such that it only occurs between the targeted node and the new $n_{in}$ adversarial nodes.
		
		\item{\textbf{META.}} META is a global attack strategy which modifies the original graph based on meta gradients. To extend the original META to the NIA scenario, we also randomly sample $n_{in}$ feature vectors and corresponding labels of original nodes as the newly injected nodes, then initialize the original neighbors following the generation of the BA scale-free model, which is the same as the extension of Nettack. After that, we employ the original META but only allow the attack operations to happen between the newly injected $n_{in}$ nodes and normal nodes. 
		
		\item{\textbf{AFGSM.}} AFGSM is a NIA method utilizing the gradients of loss functions of targeted nodes. Since AFGSM focuses on a single node attack once a time, we also extend the original AFGSM to be a global attack method like Nettack does. Specifically, for each new adversarial node, we randomly choose a node in the test set as the targeted node, and then employ the original AFGSM to achieve targeted attacks. Moreover, as AFGSM can only generate binary features, we will assign the randomly selected feature vectors of original nodes to the adversarial nodes on two datasets (i.e., Cora-ML and Pubmed) with continuous features.
		
		\item{\textbf{G-NIA.}} G-NIA is a targeted evasion NIA method by utilizing a parametric model to learn the attack patterns. Since the designs of G-NIA cannot only retain the feature consistency but also promise a given unnoticeable feature attack budget, we only need to extend it to global attacks by randomly selecting a node in the test set for each injected budget, which is the same as the change on Nettack and AFGSM.
		
		\item{\textbf{TDGIA.}} TDGIA is a well-performed global NIA method combining both the topological defective edge selections and smooth feature generations. However, the original TDGIA fails to promise a given feature-attack budget, which ignores the situation that the features of nodes may be extremely sparse vectors. Moreover, TDGIA is only suitable for generating continuous features, causing feature inconsistent issues once the target graph has binary features. Therefore, we will first assign the randomly selected feature vectors of original nodes to the adversarial nodes to ensure the imperceptible feature generations, then employ the original neighbor selections of TDGIA.
		
	\end{enumerate}

	\subsubsection{Target Models}
	For the target GNNs, we employ GANI and other baselines to attack two general GNNs including GCN \cite{kipf2017semi} and SGC \cite{wu2019simplifying}, and two defended or robust GNNs including Jaccard \cite{ijcai2019-669} and SimPGCN \cite{jin2021node}. 
	
	\begin{enumerate}
		\item{\textbf{GCN.}} GCN is one of the most representative GNNs. GCN learns the high-level representation of central nodes by aggregating the features of local neighbors layer by layer.
		
		\item{\textbf{SGC.}} SGC is the simplified version of GCN via omitting the activation functions such as ReLU in the middle layers. SGC achieves a competitive performance as original GCN while having a much smaller parameter complexity.
		
		\item{\textbf{Jaccard.}} Jaccard is a defended GNN by employing a pre-process operation on the basic GCN. Specifically, Jaccard first calculates the {\em Jaccard similarity} of connected node pairs and only preserves those links with higher similarity, based on the assumption that connected node pairs with low similarity usually will be noise. Particularly, as the features with continuous type cannot directly utilize the original Jaccard similarity, we use the Cosine similarity as a replacement on Cora-ML and Pubmed datasets.
		
		\item{\textbf{SimPGCN.}} SimPGCN is a recently proposed defended method for improving the aggregator designs of GNN. Specifically, SimPGCN will adaptively preserve both the feature similarity and structural similarity during the aggregation process to resist structural attacks. 
	\end{enumerate}
	
	Since GANI is optimized through the simple SGC model, the performance of attacking the other three GNN models indicates the generality of our method.

	
	\subsubsection{Parameter Settings} 
	Following previous work \cite{zugner2018adversarial}, the training, validation, and test ratio of all datasets are 10\%, 10\%, and 80\%, respectively. All experiments are the average performance under 5 different splits. All attack methods and attacked models are adopted from the default parameter settings. Specifically, for the proposed method, the candidate ratio $\alpha$ is set to 50\%. The population size and maximal iteration are set to 40 and 100, respectively. Moreover, the crossover rate $p_c$ and mutation rate $p_m$ are set as 0.5 and 0.3 based on the grid search process, respectively. In addition, to have a fair comparison, we employ the same link-attack budget $\Delta L$ to all baselines. It is worth noting that, Nettack and META will actually have 2 * $\Delta L$ link-attack budget because of the initialization of the BA model. Particularly, to speed up the calculation of GA in the large-scale searching space, GANI is implemented via parallel computing by employing the message passing interface \cite{gropp1999using}. The codes are publicly available at \href{https://github.com/alexfanjn/GANI}{https://github.com/alexfanjn/GANI}.

	\subsection{Experimental Analysis}
	
	\subsubsection{Overall Attack Performance on GNNs}
	
	The attack results on four GNNs of different baselines by injecting 5\% nodes are shown in Table \ref{table:gnns}. Among them, GANI achieves the best attack performance almost in all cases (with the overall best average ranking among all rows). We interestingly find that the extended methods also can obtain a remarkable attack performance on some cases, especially for Nettack. However, compared with the significant attack performance on Cora and Citeseer, these baselines perform worse on Cora-ML and Pubmed datasets, indicating that these methods may not be suitable for the datasets with continuous features. As a comparison, the proposed GANI still perform well on these types of datasets.

	Moreover, for the comparison of different GNN models, GANI obtains the best attack performance on SGC since SGC is the corresponding surrogate model, which can be considered as white-box attacks as the attackers already know the specific target model. Meanwhile, GANI also obtains a relatively good attack performance even on two defended GNNs, illustrating its strong generalization ability as GANI is trained from the surrogate SGC model.

	\begin{figure*}[htbp]
		\subfigure[\textbf{Cora}]{
			\begin{minipage}[]{0.25\linewidth}
				\includegraphics[scale=0.27]{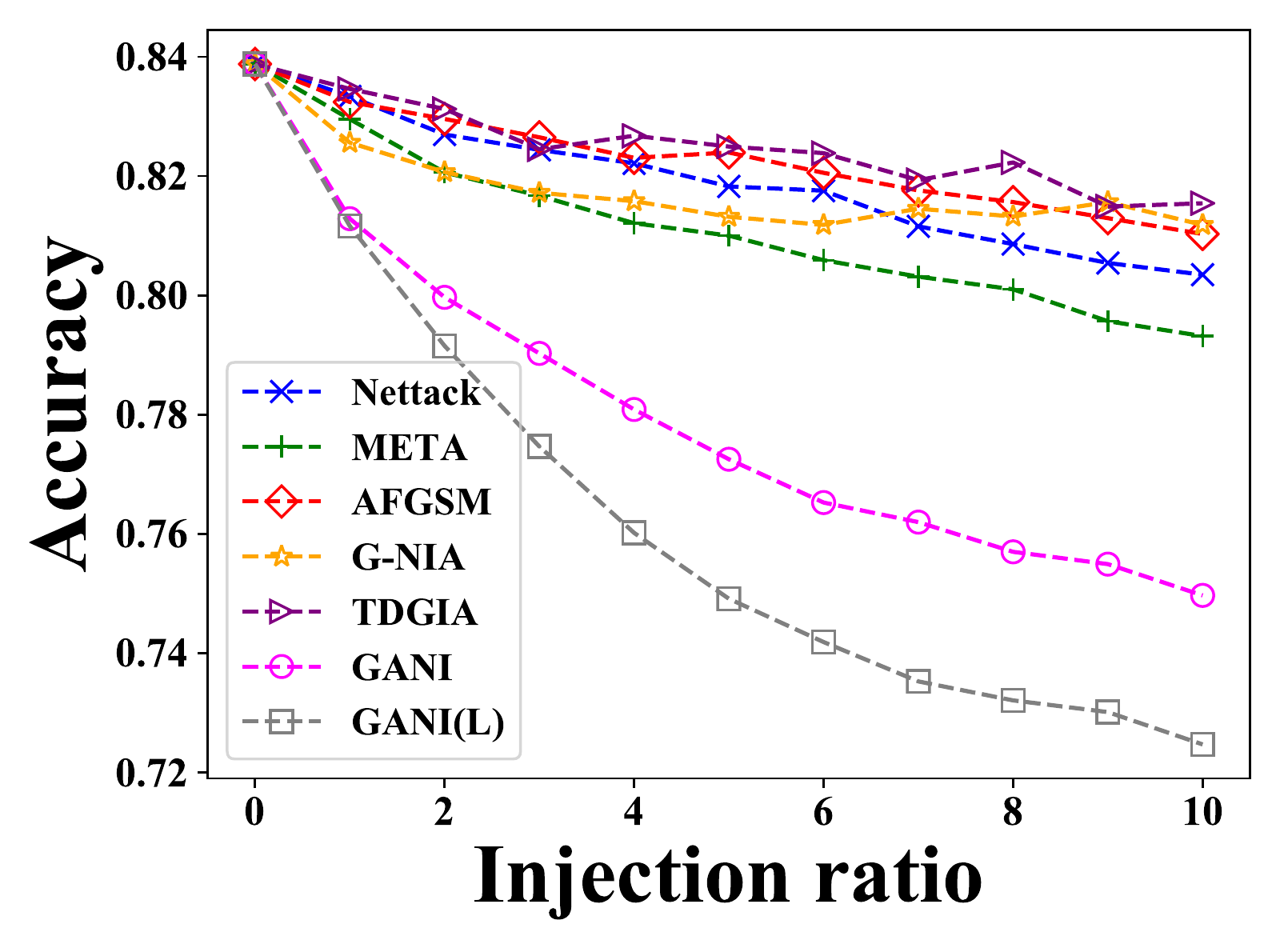}
			\end{minipage}%
		}%
		\subfigure[\textbf{Citeseer}]{
			\begin{minipage}[]{0.25\linewidth}
				\includegraphics[scale=0.27]{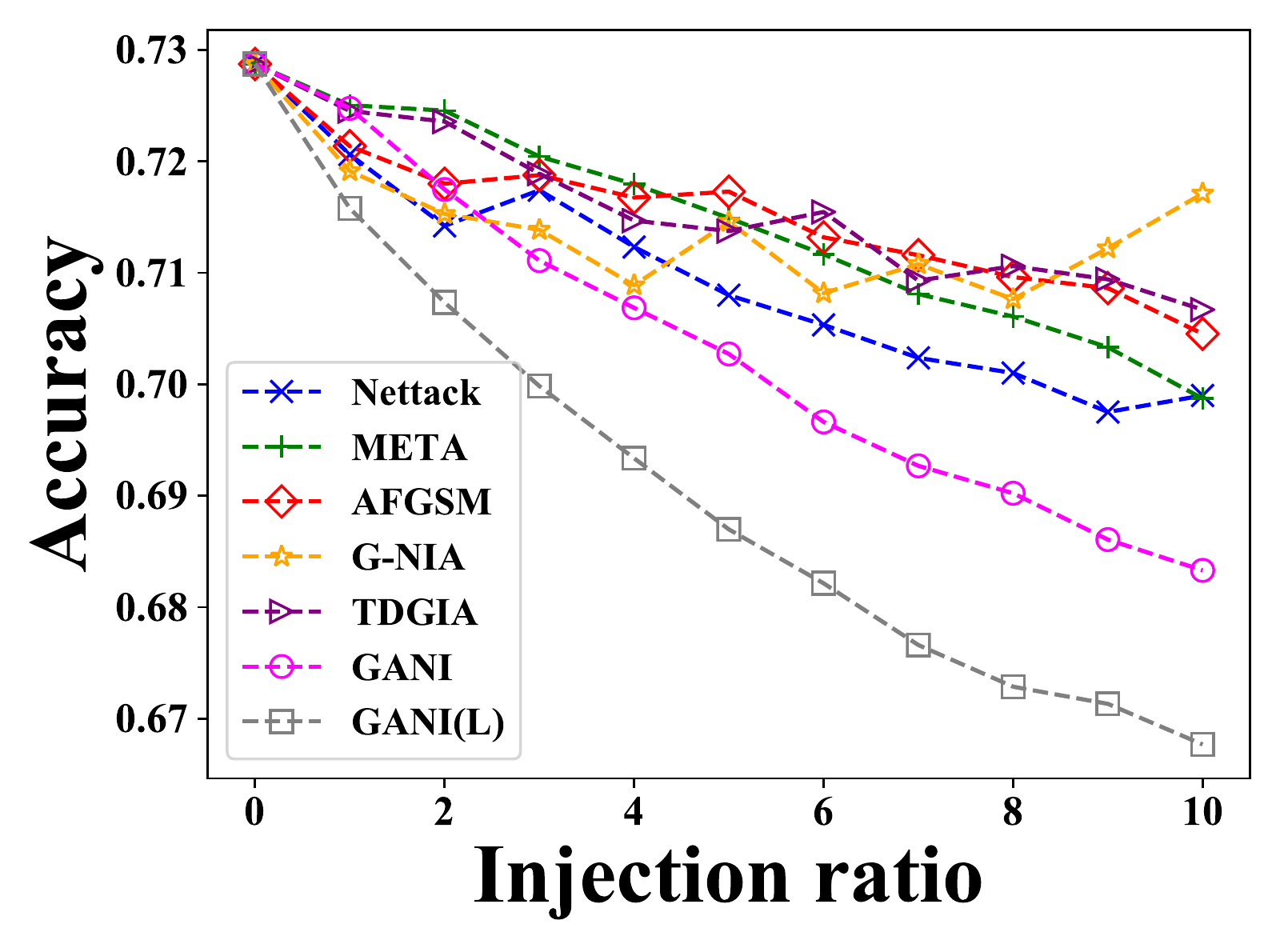}
			\end{minipage}%
		}%
		\subfigure[\textbf{Cora-ML}]{
			\begin{minipage}[]{0.25\linewidth}
				\includegraphics[scale=0.27]{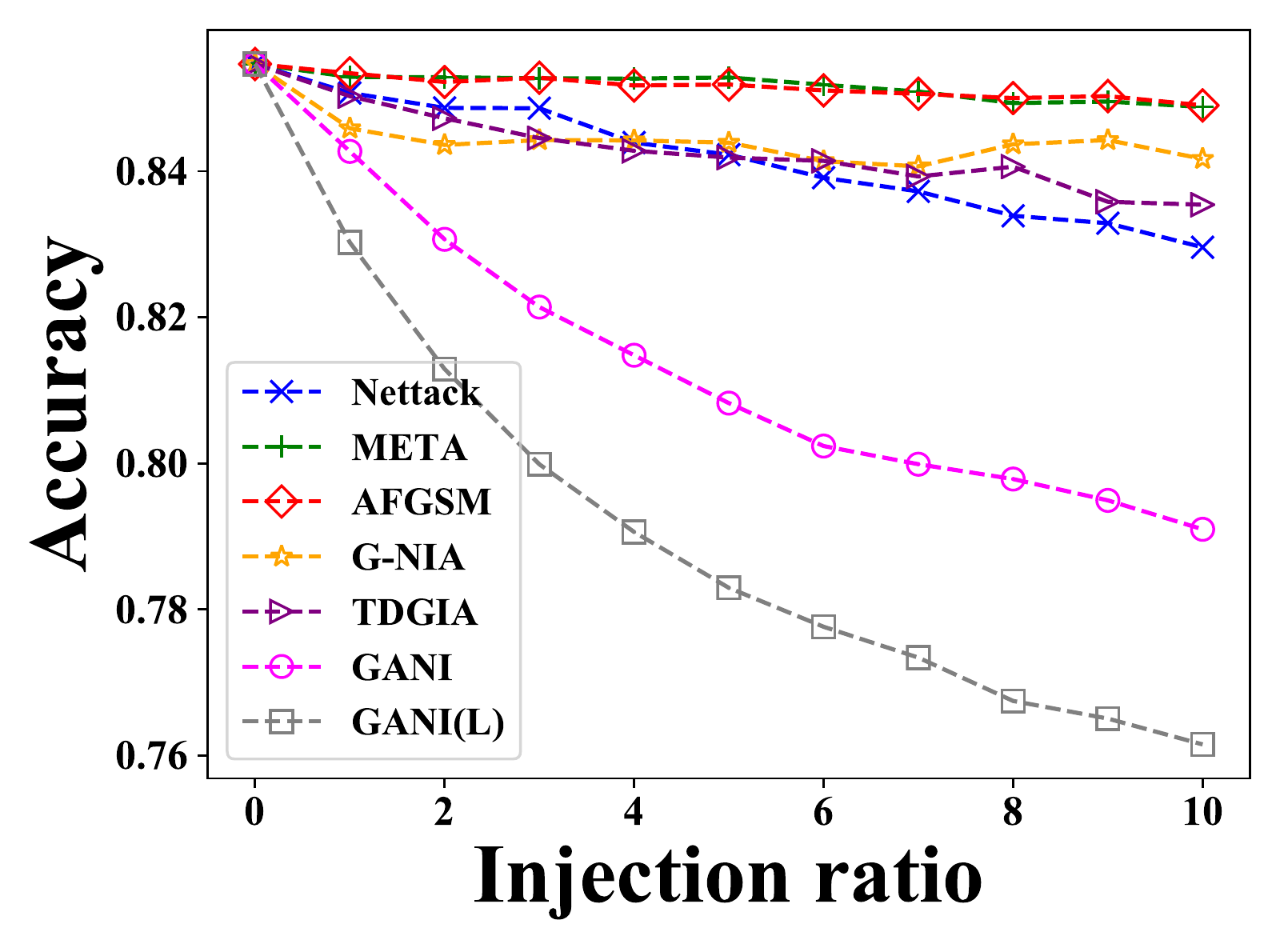}
			\end{minipage}%
		}%
		\subfigure[\textbf{Pubmed}]{
			\begin{minipage}[]{0.25\linewidth}
				\includegraphics[scale=0.27]{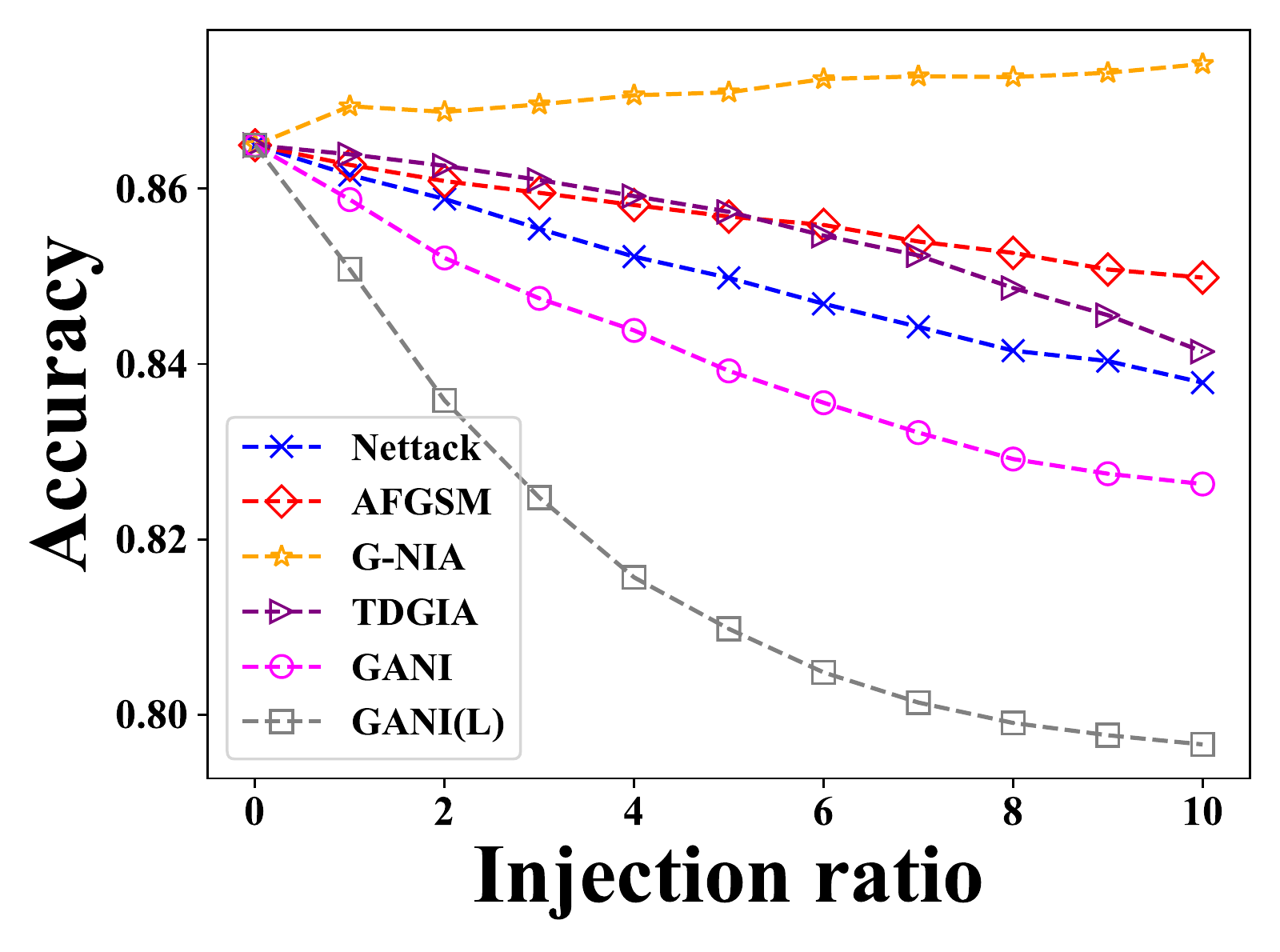}
			\end{minipage}%
		}%
		\centering
		\caption{Change  in accuracy of GCN w.r.t. different ratios of injected nodes (1\% to 10\%) in four datasets.}
		\label{fig:gcn}
	\end{figure*}

	\begin{figure*}[htbp]
		\subfigure[\textbf{Cora}]{
			\begin{minipage}[]{0.25\linewidth}
				\includegraphics[scale=0.27]{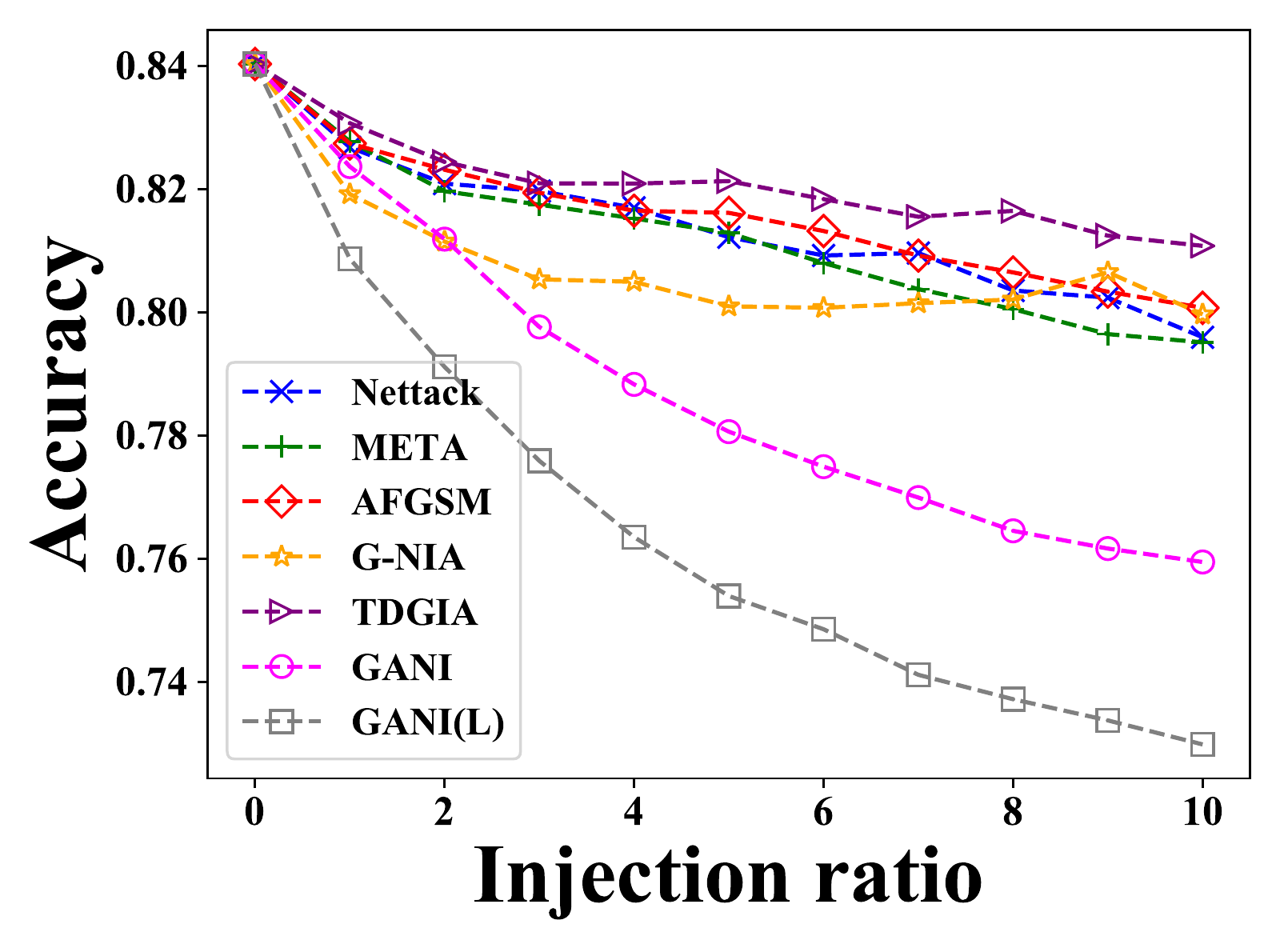}
			\end{minipage}%
		}%
		\subfigure[\textbf{Citeseer}]{
			\begin{minipage}[]{0.25\linewidth}
				\includegraphics[scale=0.27]{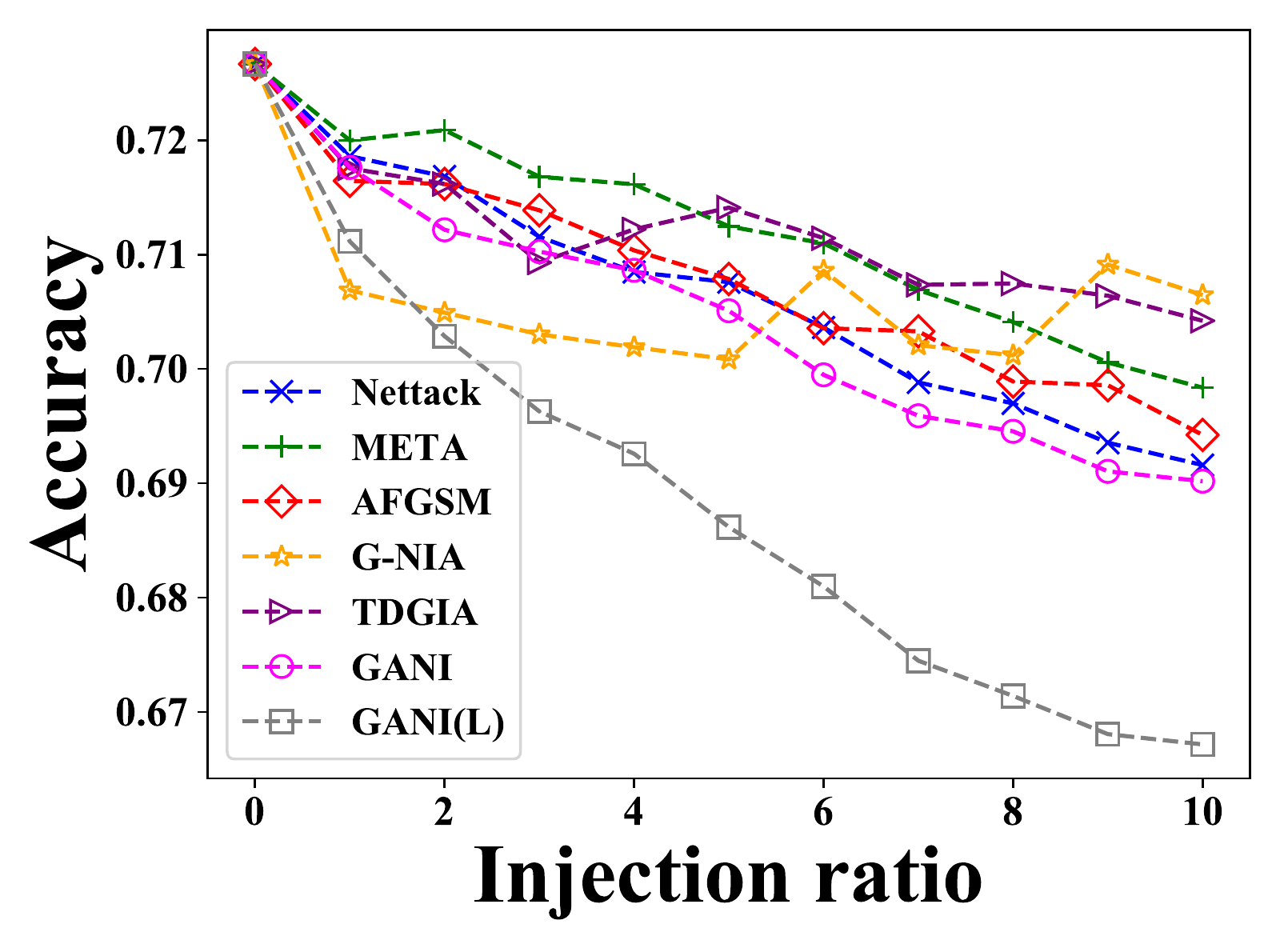}
			\end{minipage}%
		}%
		\subfigure[\textbf{Cora-ML}]{
			\begin{minipage}[]{0.25\linewidth}
				\includegraphics[scale=0.27]{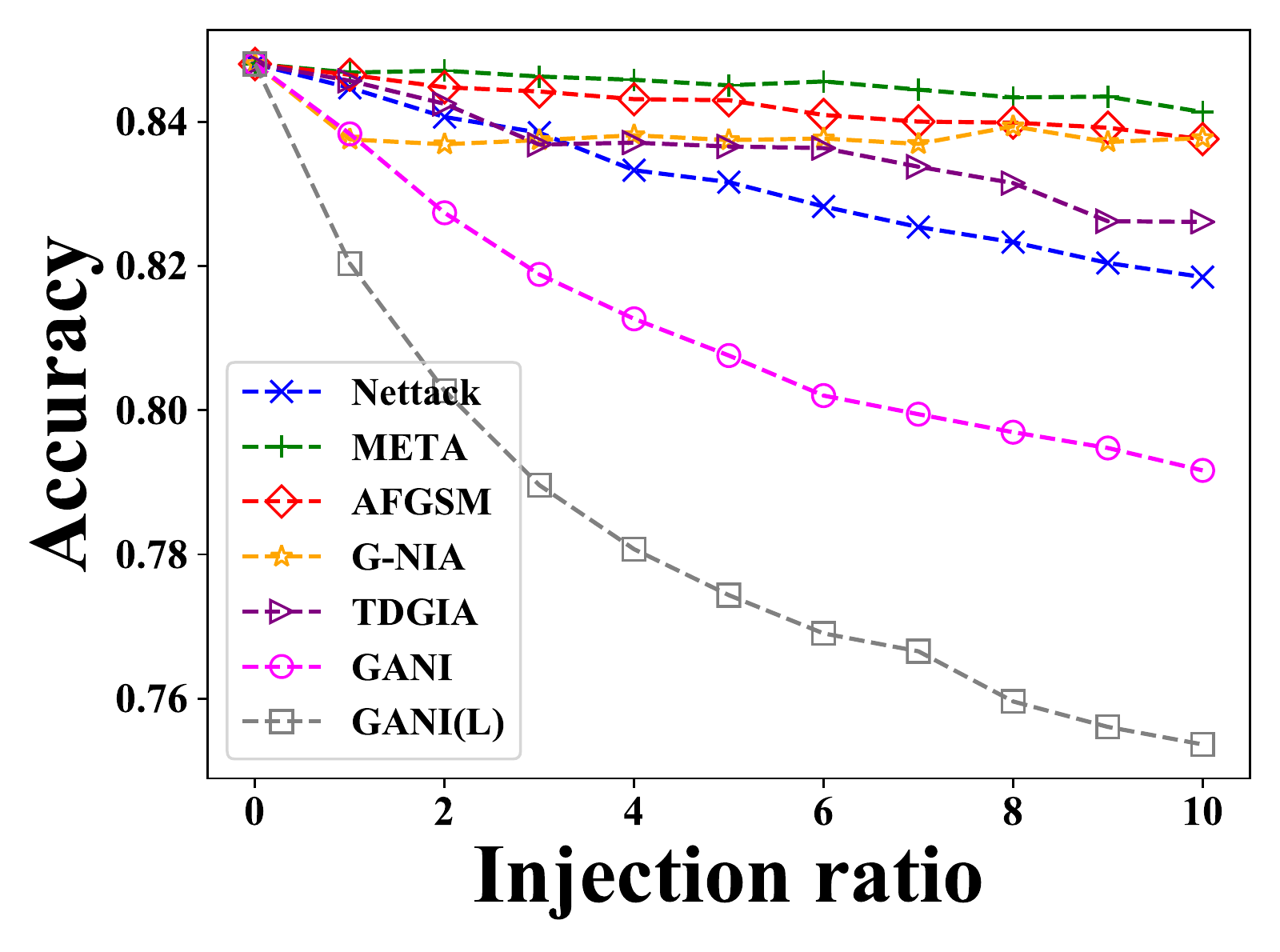}
			\end{minipage}%
		}%
		\subfigure[\textbf{Pubmed}]{
			\begin{minipage}[]{0.25\linewidth}
				\includegraphics[scale=0.27]{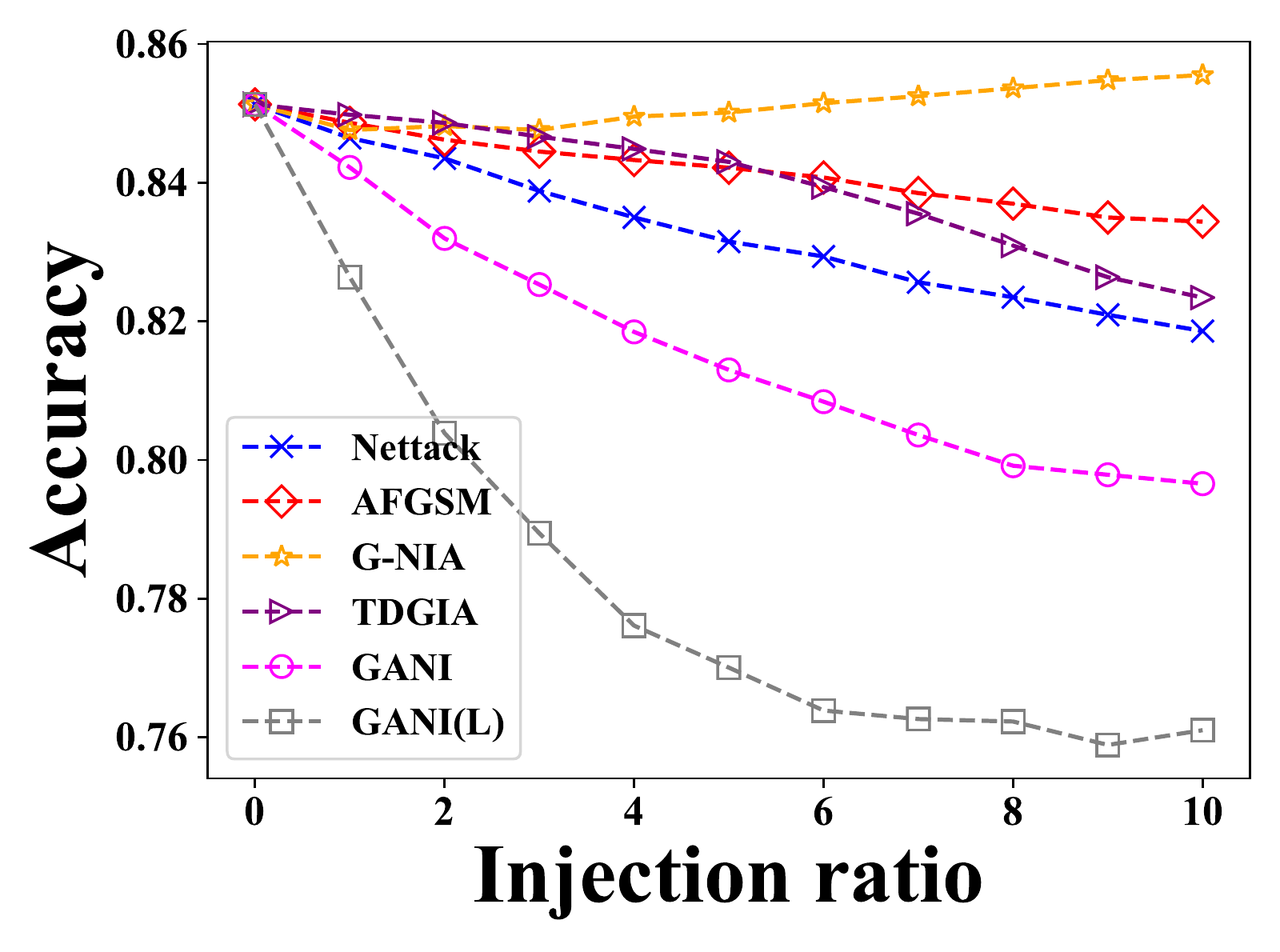}
			\end{minipage}%
		}%
		\centering
		\caption{Change  in accuracy of SGC w.r.t. different ratios of injected nodes (1\% to 10\%) in four datasets.}
		\label{fig:sgc}
	\end{figure*}

	\begin{figure*}[htbp]
		\subfigure[\textbf{Cora}]{
			\begin{minipage}[]{0.25\linewidth}
				\includegraphics[scale=0.27]{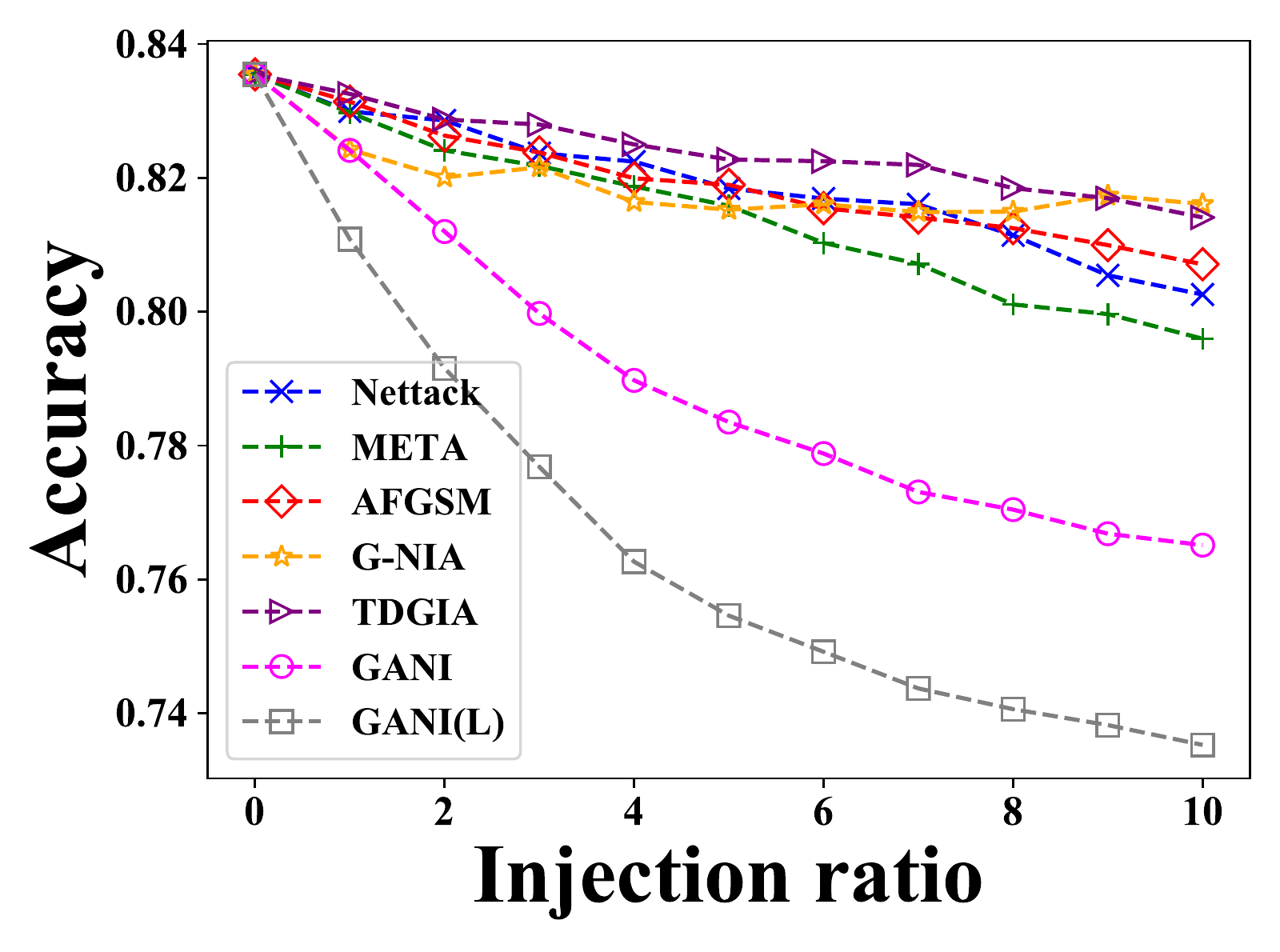}
			\end{minipage}%
		}%
		\subfigure[\textbf{Citeseer}]{
			\begin{minipage}[]{0.25\linewidth}
				\includegraphics[scale=0.27]{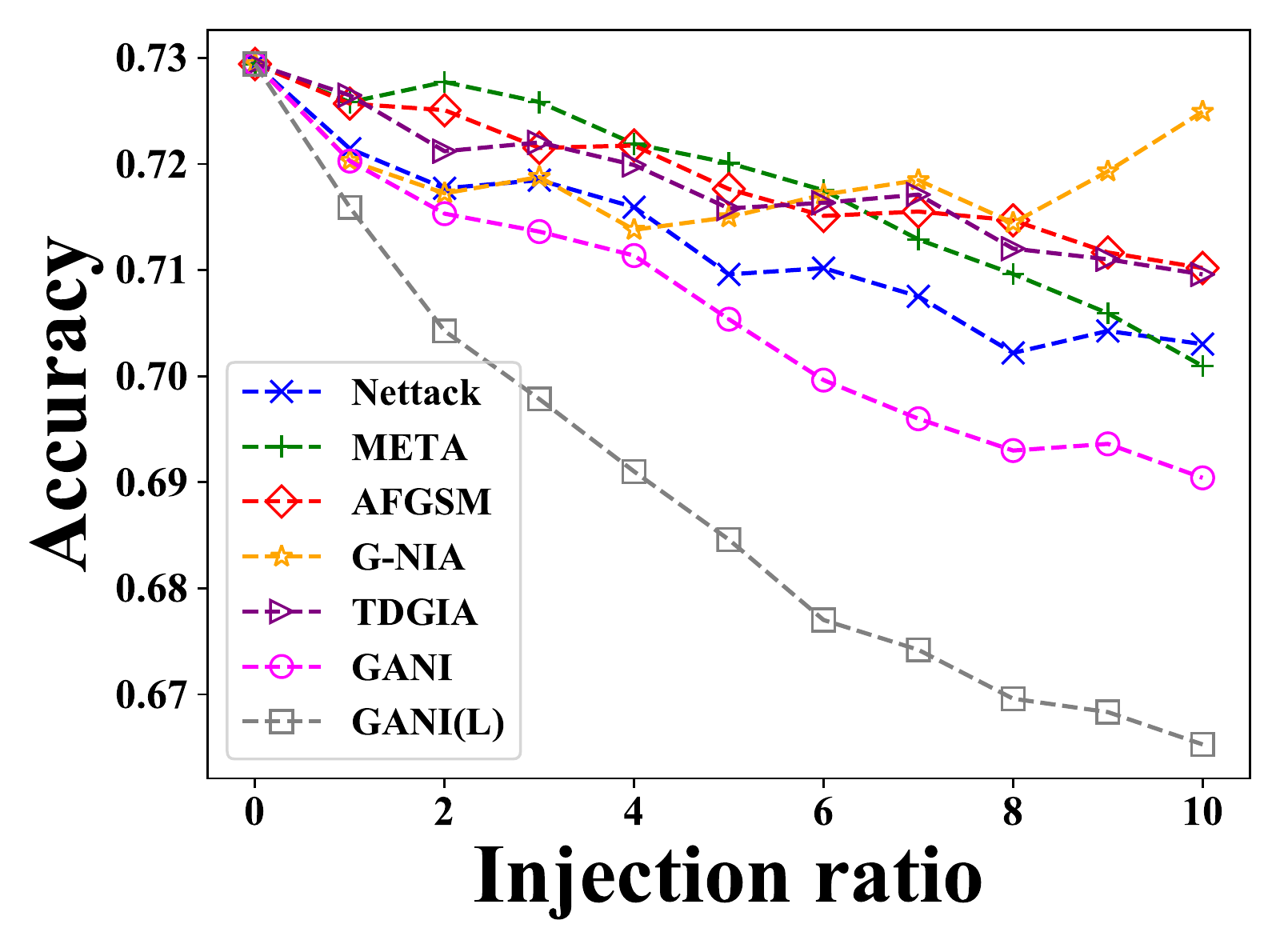}
			\end{minipage}%
		}%
		\subfigure[\textbf{Cora-ML}]{
			\begin{minipage}[]{0.25\linewidth}
				\includegraphics[scale=0.27]{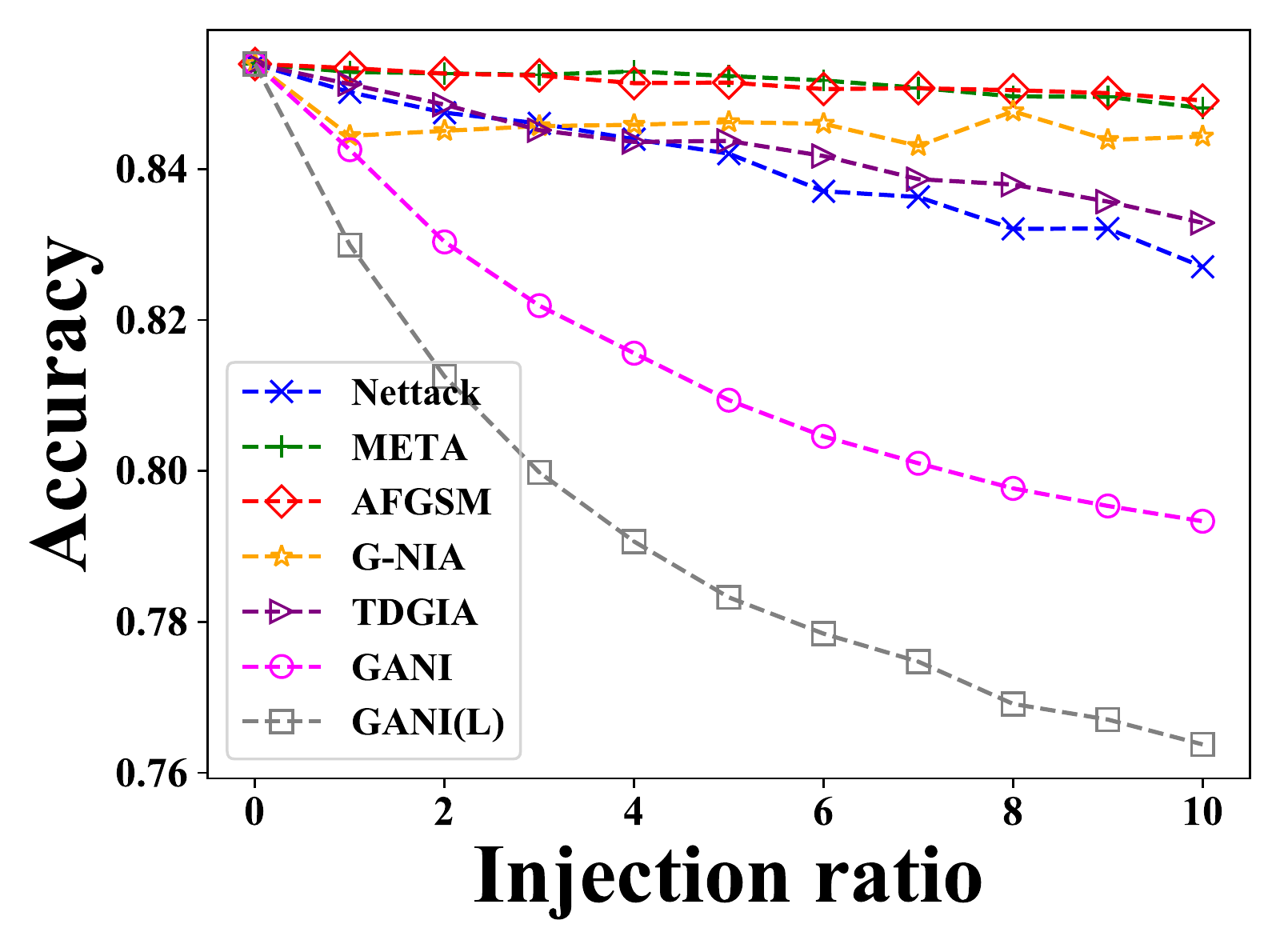}
			\end{minipage}%
		}%
		\subfigure[\textbf{Pubmed}]{
			\begin{minipage}[]{0.25\linewidth}
				\includegraphics[scale=0.27]{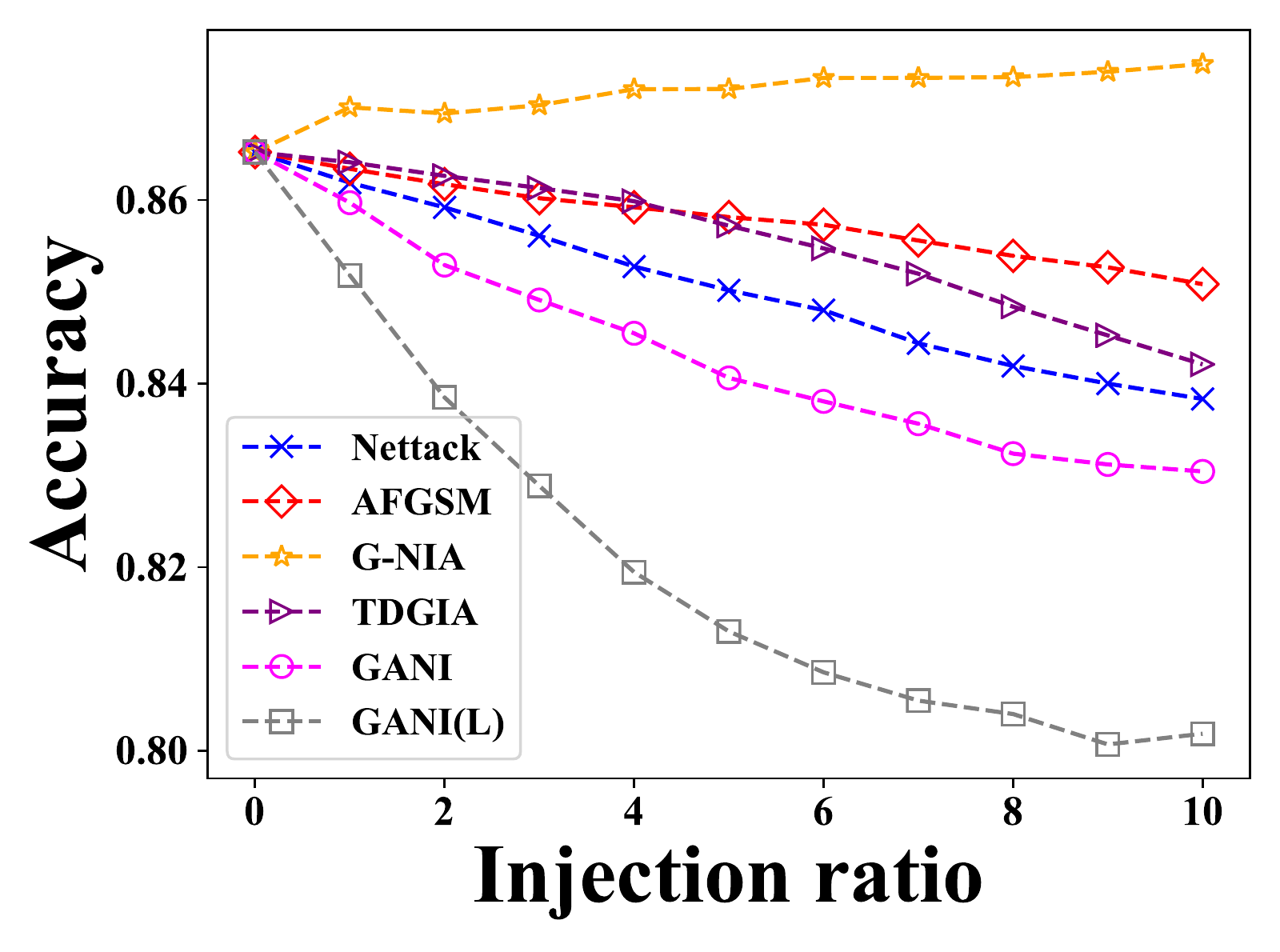}
			\end{minipage}%
		}%
		\centering
		\caption{Change in accuracy of Jaccard w.r.t. different ratios of injected nodes (1\% to 10\%) in four datasets.}
		\label{fig:jaccard}
	\end{figure*}

	\begin{figure*}[htbp]
		\subfigure[\textbf{Cora}]{
			\begin{minipage}[]{0.25\linewidth}
				\includegraphics[scale=0.27]{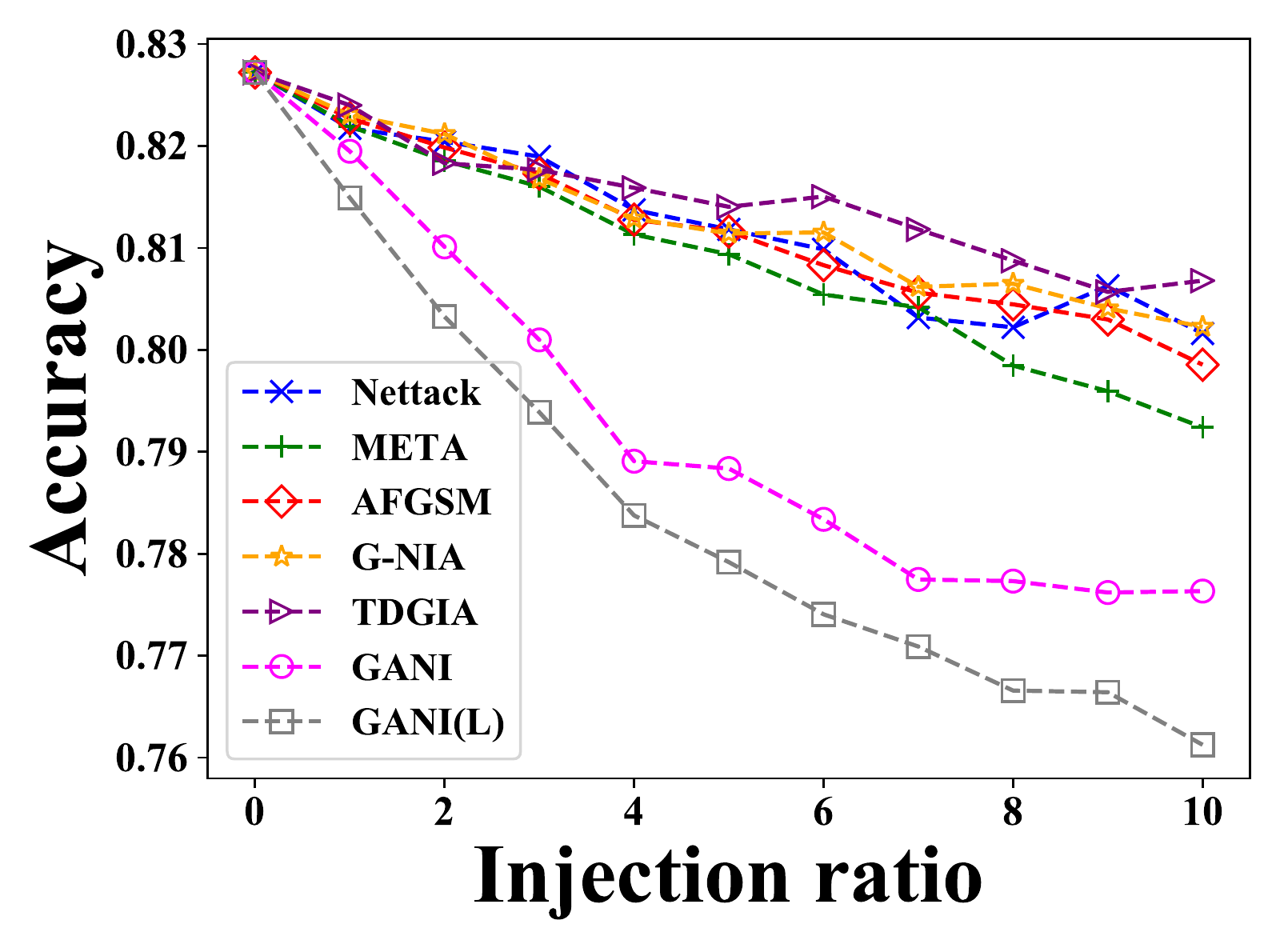}
			\end{minipage}%
		}%
		\subfigure[\textbf{Citeseer}]{
			\begin{minipage}[]{0.25\linewidth}
				\includegraphics[scale=0.27]{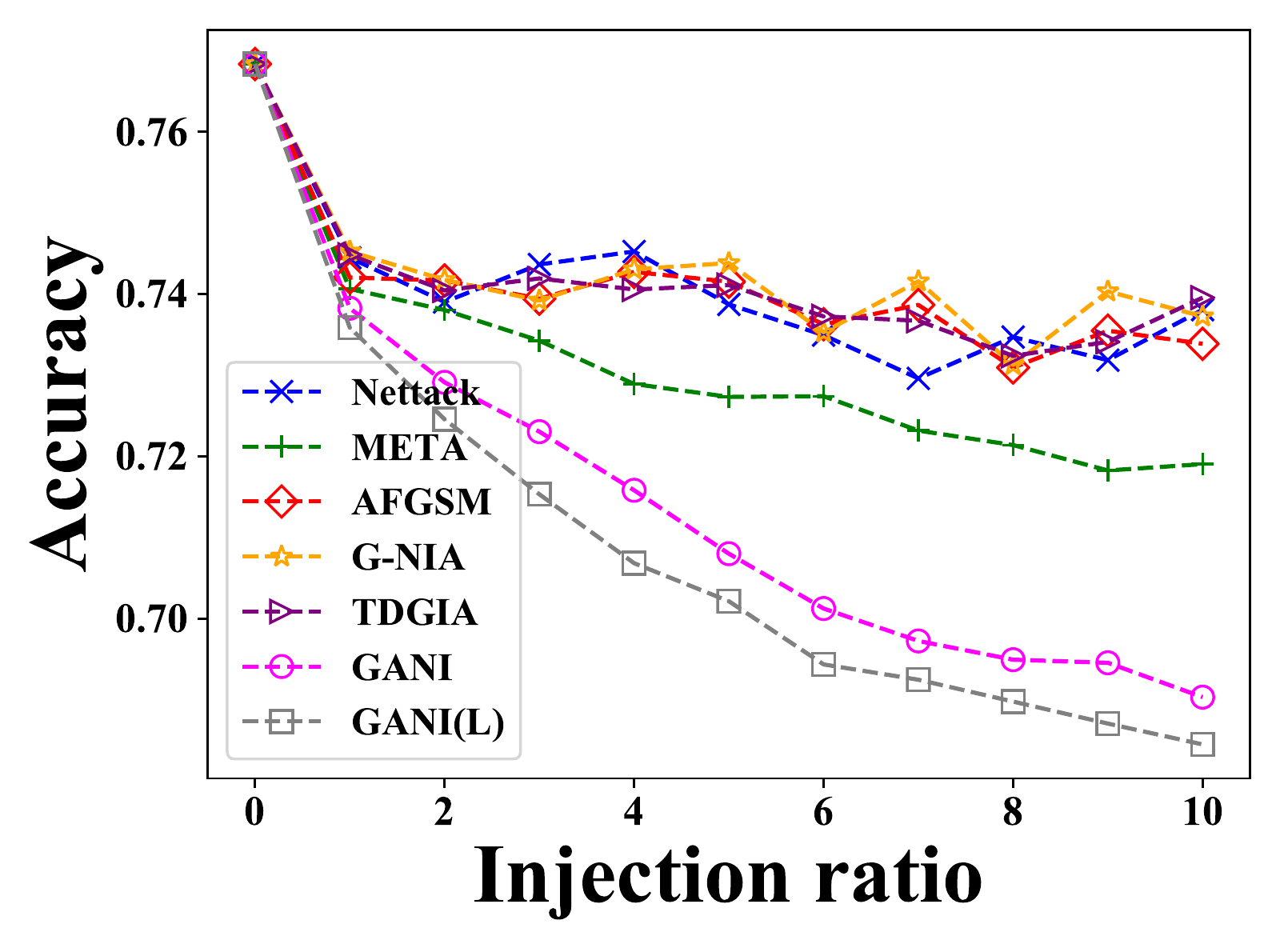}
			\end{minipage}%
		}%
		\subfigure[\textbf{Cora-ML}]{
			\begin{minipage}[]{0.25\linewidth}
				\includegraphics[scale=0.27]{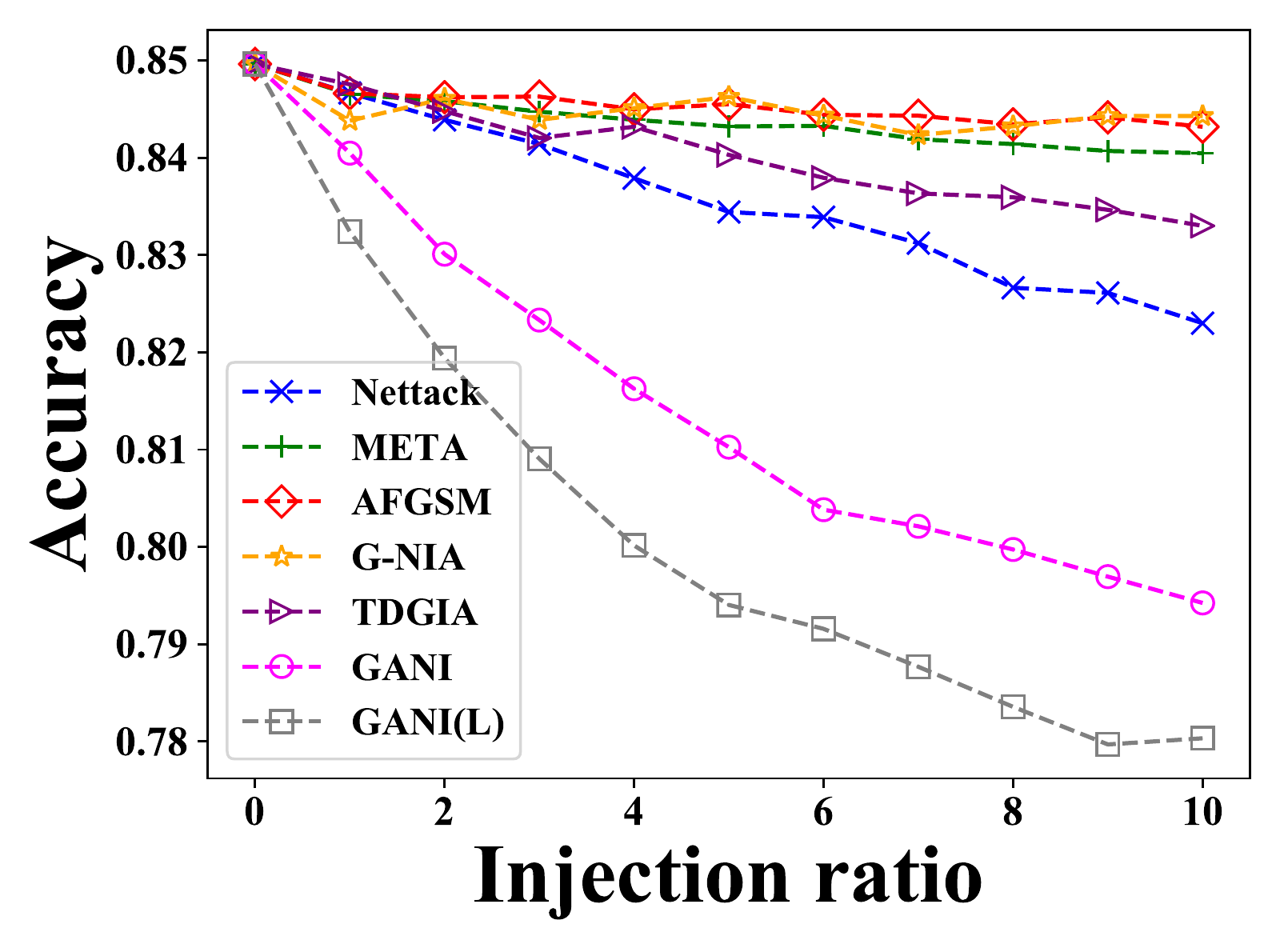}
			\end{minipage}%
		}%
		\subfigure[\textbf{Pubmed}]{
			\begin{minipage}[]{0.25\linewidth}
				\includegraphics[scale=0.27]{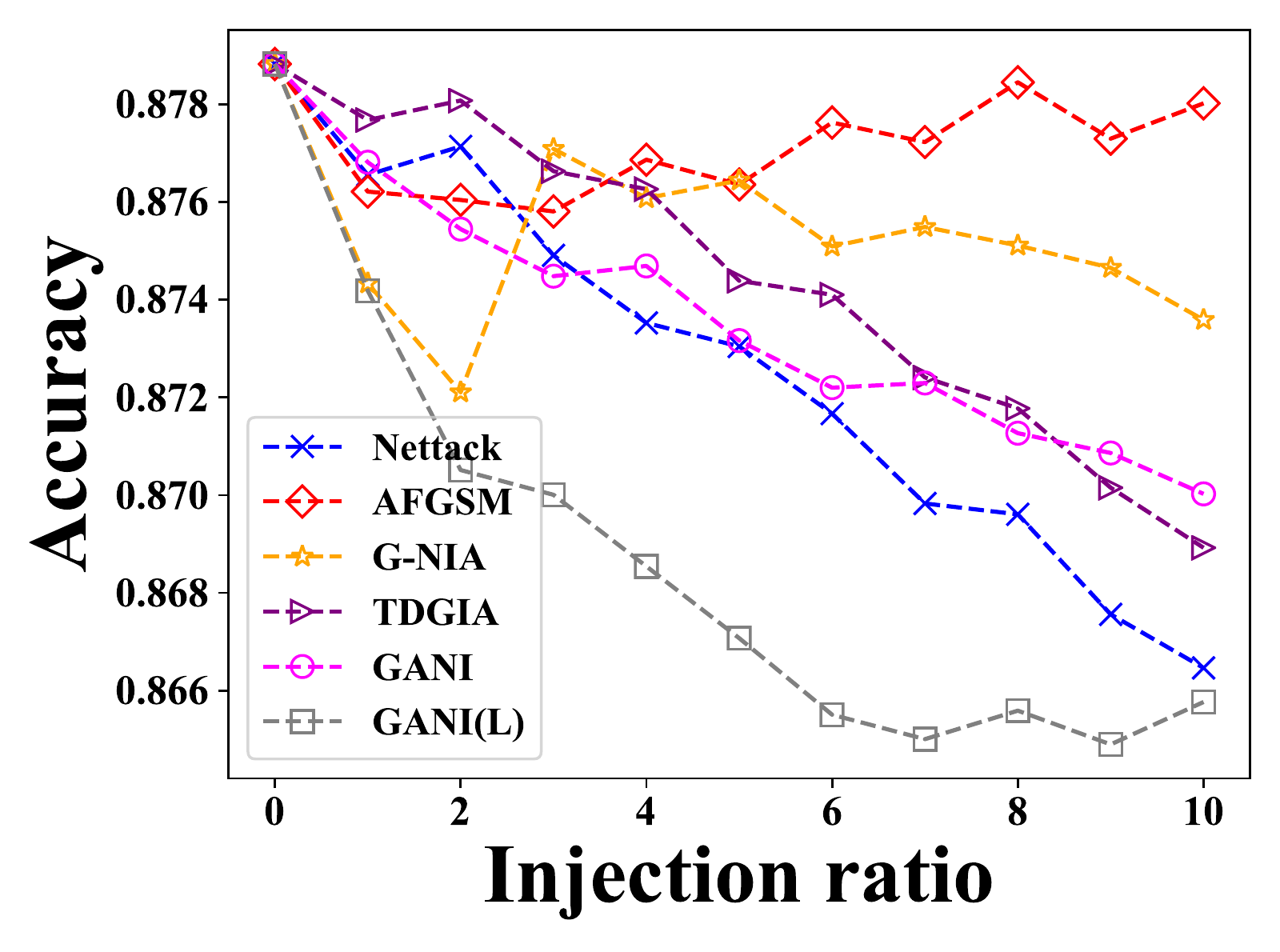}
			\end{minipage}%
		}%
		\centering
		\caption{Change in accuracy of SimPGCN w.r.t. different ratios of injected nodes (1\% to 10\%) in four datasets.}
		\label{fig:simpgcn}
	\end{figure*}

	\begin{figure*}[htbp]
		\subfigure[\textbf{Cora}]{
			\begin{minipage}[]{0.25\linewidth}
				\includegraphics[scale=0.27]{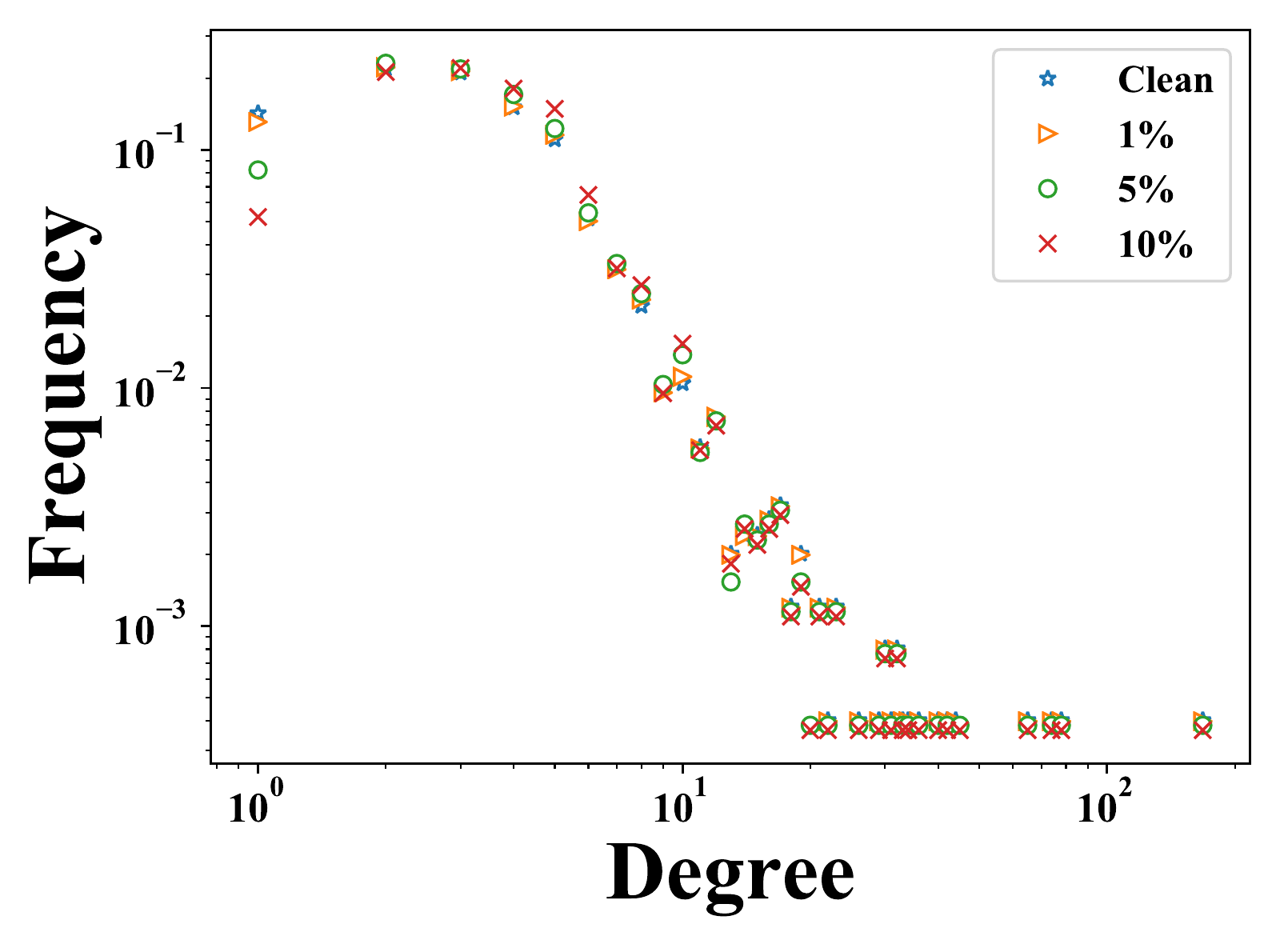}
			\end{minipage}%
		}%
		\subfigure[\textbf{Citeseer}]{
			\begin{minipage}[]{0.25\linewidth}
				\includegraphics[scale=0.27]{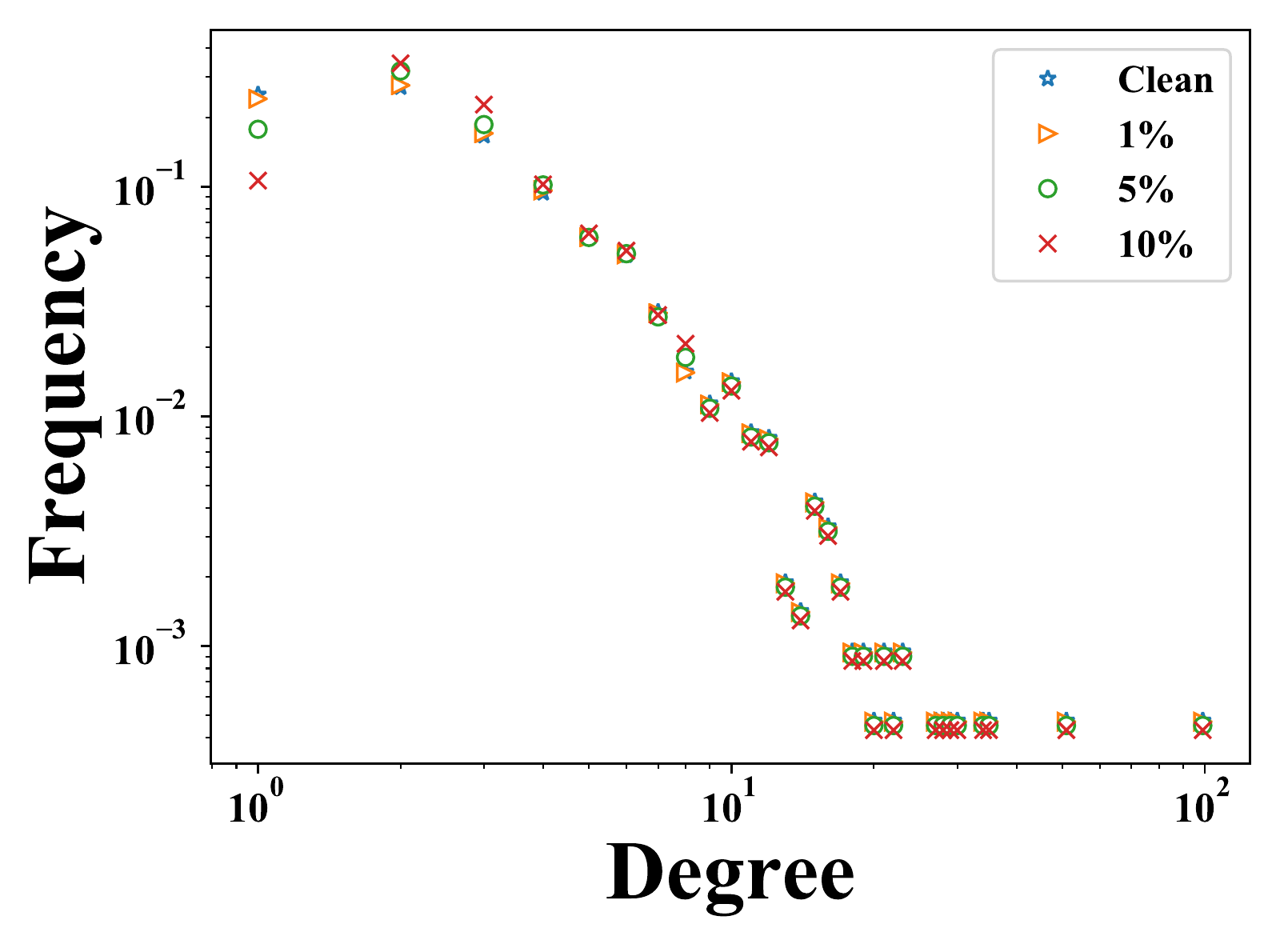}
			\end{minipage}%
		}%
		\subfigure[\textbf{Cora-ML}]{
			\begin{minipage}[]{0.25\linewidth}
				\includegraphics[scale=0.27]{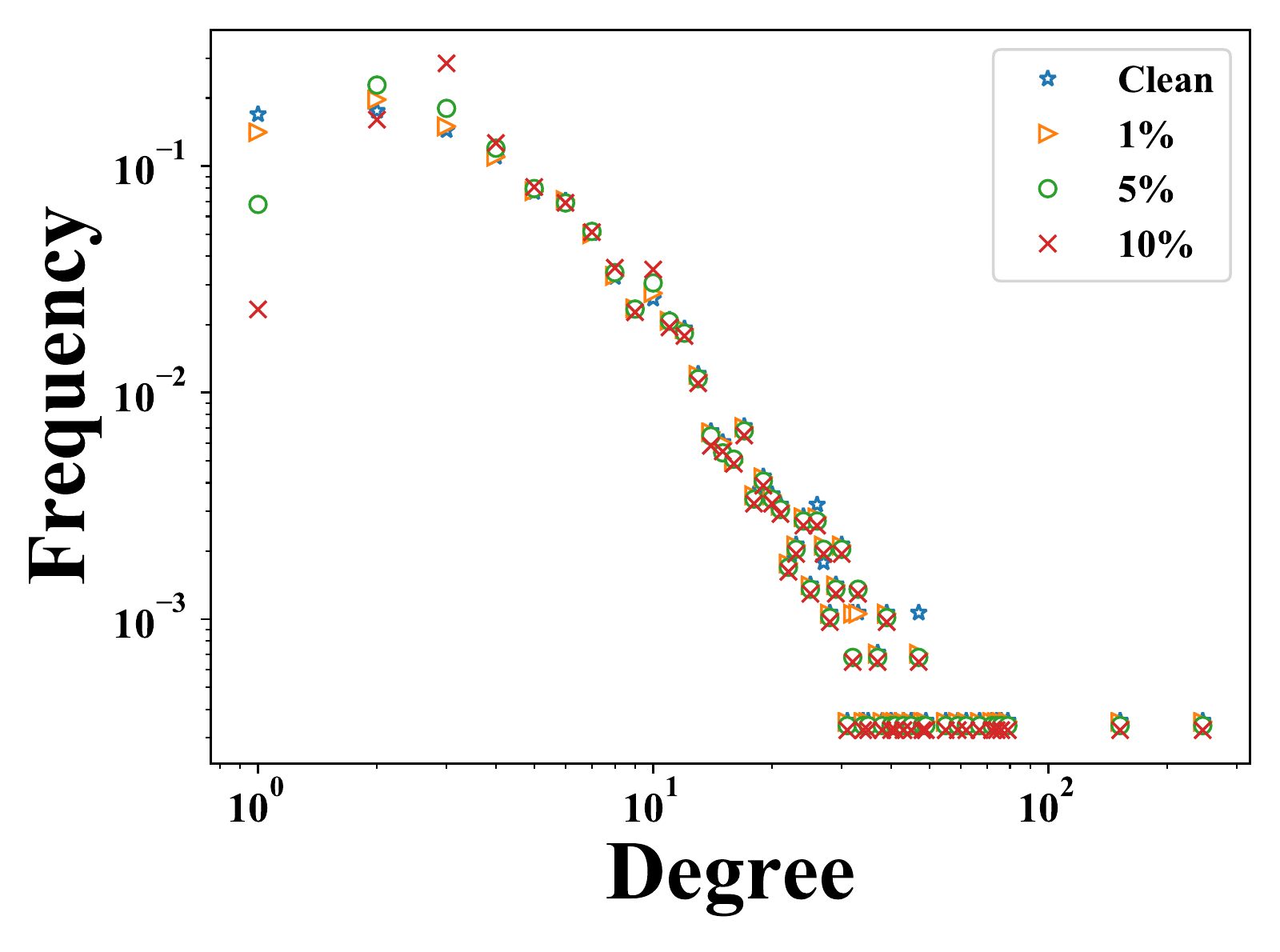}
			\end{minipage}%
		}%
		\subfigure[\textbf{Pubmed}]{
			\begin{minipage}[]{0.25\linewidth}
				\includegraphics[scale=0.27]{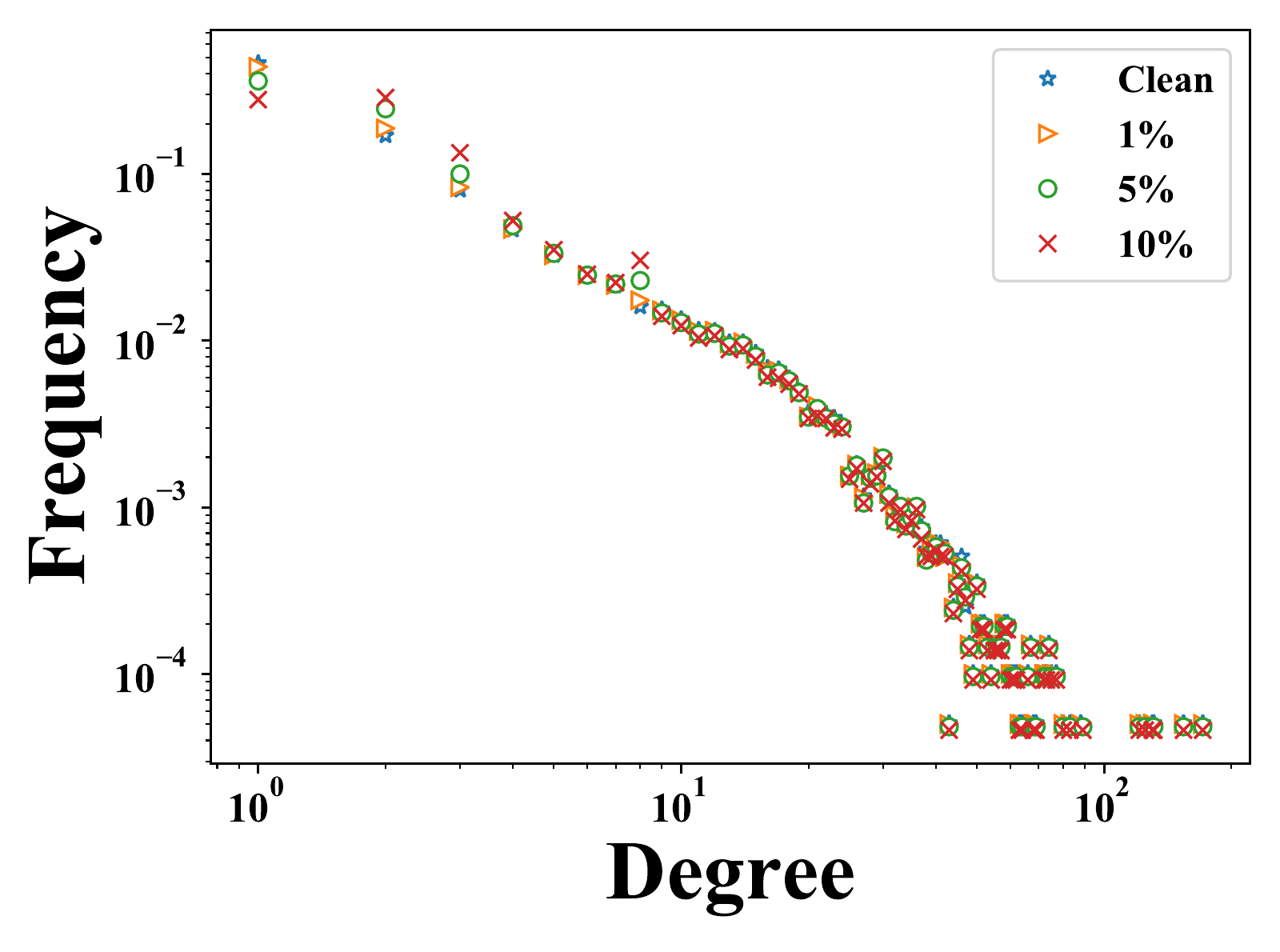}
			\end{minipage}%
		}%
		\centering
		\caption{Change in degree distribution under different injection ratios (i.e., 1\%, 5\% and 10\%) in four datasets.}
		\label{fig:degree_properties}
	\end{figure*}

	\begin{figure*}[htbp]
		\subfigure[\textbf{Cora}]{
			\begin{minipage}[]{0.25\linewidth}
				\includegraphics[scale=0.27]{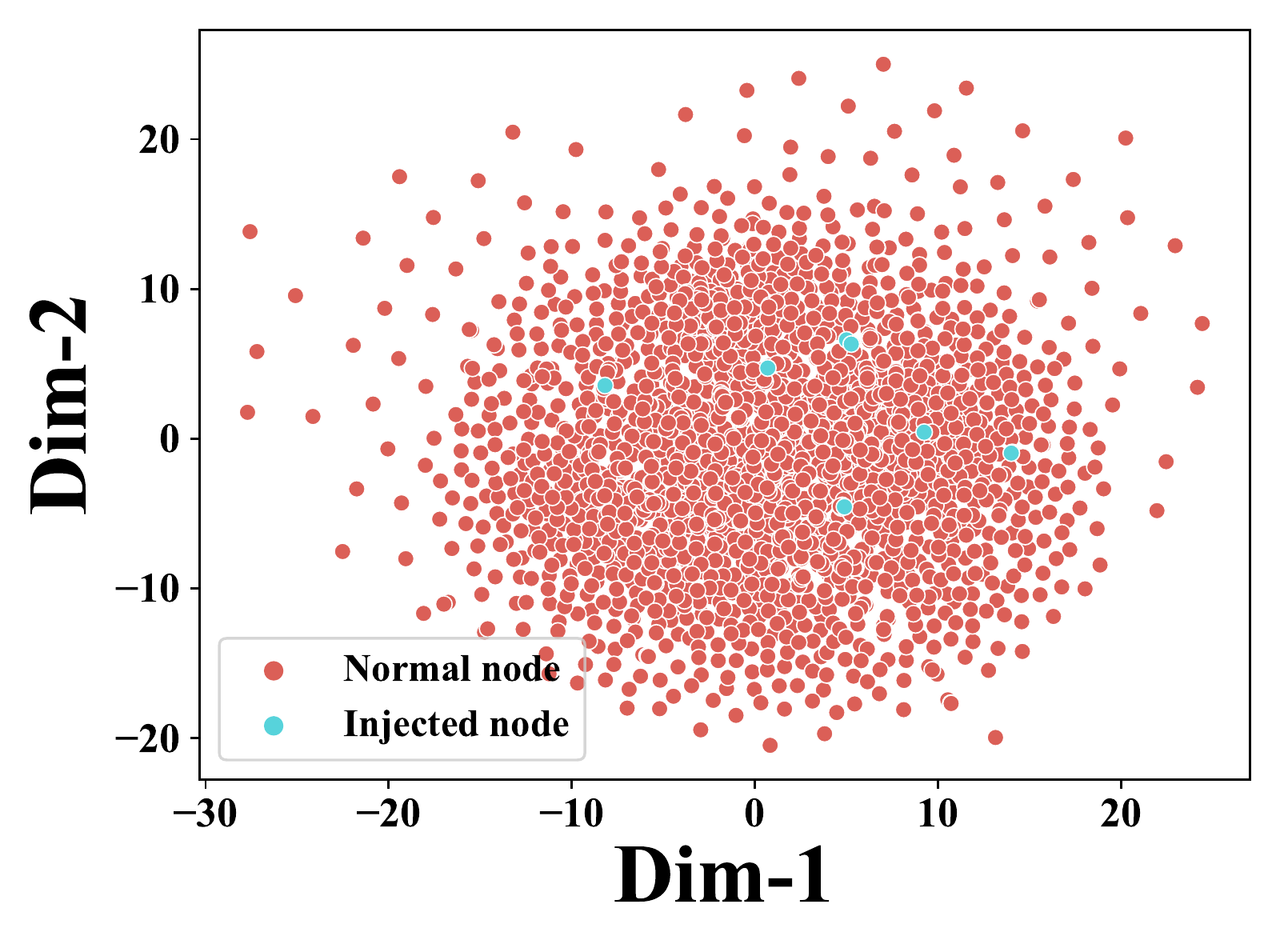}
			\end{minipage}%
		}%
		\subfigure[\textbf{Citeseer}]{
			\begin{minipage}[]{0.25\linewidth}
				\includegraphics[scale=0.27]{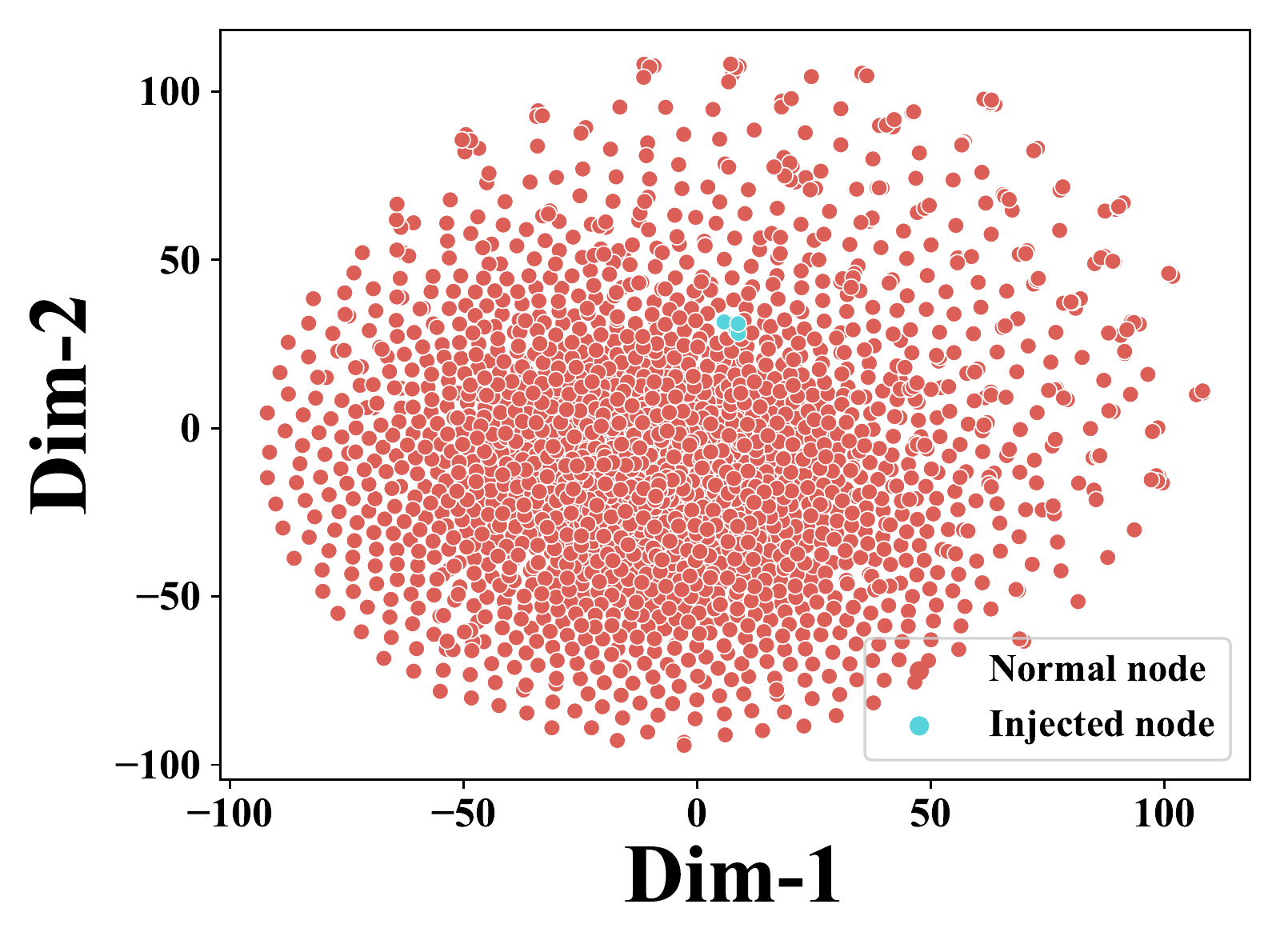}
			\end{minipage}%
		}%
		\subfigure[\textbf{Cora-ML}]{
			\begin{minipage}[]{0.25\linewidth}
				\includegraphics[scale=0.27]{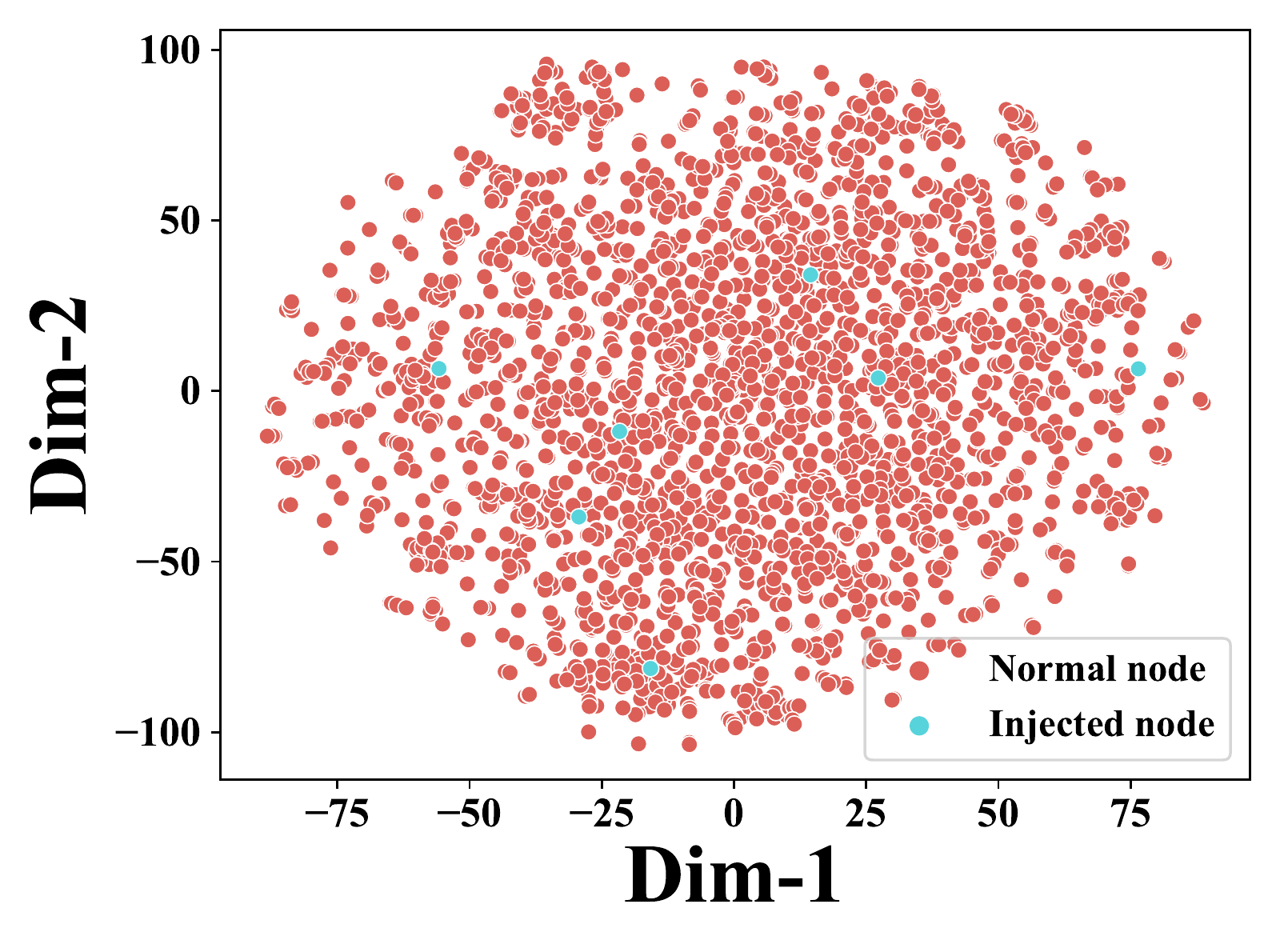}
			\end{minipage}%
		}%
		\subfigure[\textbf{Pubmed}]{
			\begin{minipage}[]{0.25\linewidth}
				\includegraphics[scale=0.27]{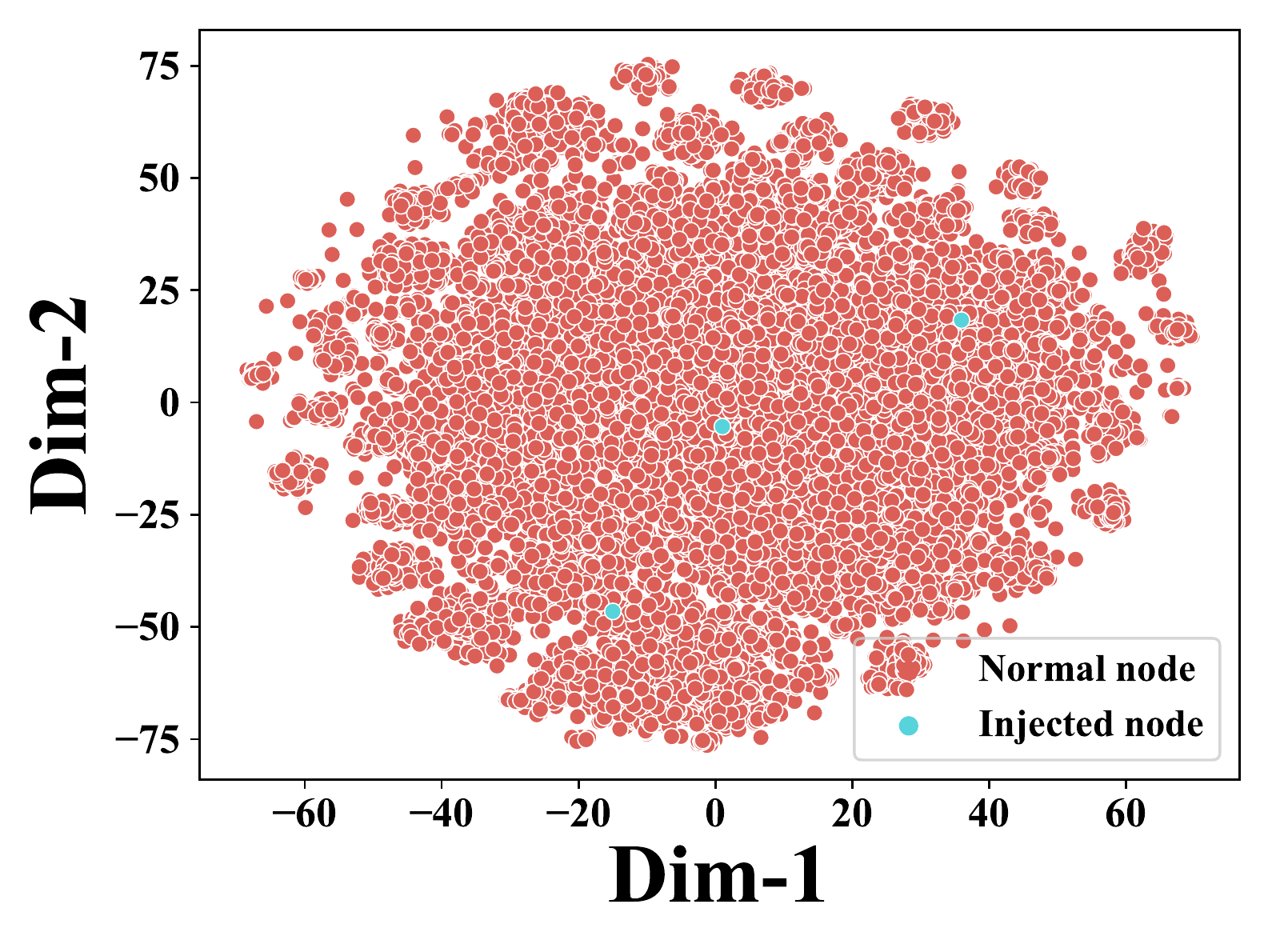}
			\end{minipage}%
		}%
		\centering
		\caption{t-SNE visualization of original features and injected features generated from GANI in four datasets.}
		\label{fig:feature_properties}
	\end{figure*}

	\subsubsection{Attack Performance w.r.t Injection Ratio}
	
	Figs. \ref{fig:gcn}, \ref{fig:sgc}, \ref{fig:jaccard}, and \ref{fig:simpgcn} are the accuracy change on four attacked GNNs by improving the injected ratio from 1\% to 10\%. Taking Fig. \ref{fig:gcn} which illustrates the attack performance on GCN as an example, as expected, the test accuracy will continuously decrease when more nodes have been injected under the guides of GANI. Some of the other baselines cannot perform well even when the injected ratio is large, especially on those datasets with continuous features. Since Nettack, G-NIA and AFGSM are extended from targeted attack scenarios, it is possible that they may not suitable for global attacks considered in our problem. For instance, the classification performance of GCN on the perturbed graph after we adopted G-NIA is even worse than the classification performance on the clean graph. Moreover, the proposed GANI can obtain the best attack performance almost in all injection ratios, which is consistent with the above analysis.
	
	In addition, Figs. \ref{fig:sgc}, \ref{fig:jaccard}, and \ref{fig:simpgcn} also draw a similar conclusion as above. The proposed GANI can achieve an overall better attack performance almost in all cases in the remaining three target GNNs. Though NETTACK and G-NIA perform slightly better than GANI in a few cases, their performance fluctuates on a large scale among different GNNs and datasets.

	\subsubsection{With or Without Label Information}
	In the default setting of GANI, we only utilize the predicted label obtained from surrogate SGC model $M_s$ to provide the label information. We want to verify whether the ground truth label can help improve the attack performance. Therefore, we replace the predicted label of SGC with the ground truth label on the optimization of GANI. By adopting this change, our GANI can obtain correct label information on feature generations, calculation of DNH, and elite selections. Based on the results (labeled as GANI(L)) from the last column in Table \ref{table:gnns} and Figs. \ref{fig:gcn} to \ref{fig:simpgcn}, we know that the ground truth label information can largely improve the attack performance as it supplies a 100\% correct guideline for GANI. This also indicates that the label information plays a critical role in the attack procedure. The proposed method performs better if there are more accurately labeled data.

	\subsubsection{Unnoticeable Perturbations}
	
	Then, we analyze whether the adversarial graphs obtained from the proposed GANI are unnoticeable from both topology and feature aspects. Specifically, we plot the degree distribution under different injection ratios and the t-SNE \cite{van2008visualizing} visualization of features. As shown in Fig. \ref{fig:degree_properties}, the degree distributions of adversarial graphs after injecting malicious nodes under different ratios are similar to the degree distributions of clean graphs. Moreover, the adversarial graphs have a lower frequency for lower degree nodes than the clean graphs, showing that GANI prefers to connect the newly injected node with those nodes having a relatively lower degree. The above finding is also consistent with the fact that it is harder to influence the representation or classification of high-degree nodes than low-degree nodes.
	
	Fig. \ref{fig:feature_properties} shows the corresponding t-SNE visualization comparison between the generated features and original features. Since our GANI generates the adversarial features based on the statistical feature information of different classes, the adversarial nodes with the same assigned label will have the same features. Therefore, the number of injected nodes in Fig. \ref{fig:feature_properties} represents the number of classes. We can observe that the generated features are projected into a similar latent place as the original features, rather than some outliers.
	
	Besides testing the imperceptibility of GANI in terms of both degree distribution and feature visualization, we also want to investigate whether the injected nodes generated from GANI are detectable when a detection algorithm is employed. Therefore, we further verify the imperceptibility of GANI from the perspective of graph outlier detection \cite{9565320}. Specifically, we employ a strong outlier detection algorithm, DOMINANT \cite{ding2019deep}, as the detector, to evaluate whether the proposed GANI is imperceptible enough. DOMINANT is a GCN-based outlier detection method which can not only detect the topological anomaly, but also identify the potential feature outliers at the same time. DOMINANT will first learn the low-dimensional representations of nodes by using GCN as an encoder. Then, the corresponding decoders are proposed to reconstruct the structure and feature information. The corresponding reconstruction error of nodes will be utilized to label the possible outliers.
	
	In this experiment, we first assume that all newly injected nodes in the perturbed graph are outliers. Then, we input the perturbed graph to DOMINANT for detection. The three commonly used evaluation metrics in outlier detection are as follows.
	
	\begin{enumerate}
		\item AUC. AUC is a metric in outlier detection methods to characterize the area under ROC (i.e., receiver operating characteristic) curve. A higher AUC indicates a better detection performance. Generally, we consider the detector performs well if AUC is close to 1. 
		\item Precision@$K$. Upon completion of the outlier detection, a rank list corresponding to the anomalous scores of nodes will be generated. Precision@$K$ is a metric to measure the ratio between the number of true outliers that are detected and the number of outliers that the detector predicted, i.e., $K$.
		\item Recall@$K$. Unlike Precision@$K$, Recall@$K$ is used to count the true outliers among the detected total true outliers (i.e., $K$).  
	\end{enumerate}

	Particularly, as we want to investigate whether the newly injected nodes will be detected by DOMINANT, we set $K$ as the total number of injected nodes, namely 5\% of original nodes. 
	
	Table \ref{table:od} is the detection performance of the perturbed graph generated from GANI. We can observe that the AUCs that DOMINANT obtained are all smaller than 0.5. Moreover, all the Precision@$K$ and Recall@$K$ are equal to 0, indicating that none of the injected nodes will be regarded as outliers. In other words, the injected nodes follow the normal patterns of the original nodes on both topological structure and feature aspects.  
	
	\begin{table}[]
		\centering
		\renewcommand\arraystretch{1.1}
		\caption{AUC, Precision@$K$, and Recall@$K$ of DOMINANT method with injecting 5\% nodes.}
		\begin{tabular}{c | c c c c}
			\bottomrule
			\textbf{Metrics}         & \textbf{Cora} & \textbf{Citeseer} & \textbf{Cora-ML} & \textbf{Pubmed}
			
			\\ \bottomrule\bottomrule
			AUC & 0.3231 & 0.4304 & 0.4920 & 0.1979
			\\ 
			Precision@$K$ & 0 & 0  & 0 & 0 \\ 
			Recall@$K$ & 0 & 0  & 0 & 0 \\
			\bottomrule
		\end{tabular}
		
		\label{table:od}
	\end{table}
	
	To sum up, the above finding indicates that the degree sampling and feature generation mechanisms of the proposed GANI can retain the topological properties and feature distributions of the clean graph, and outlier detection methods could hardly detect the injected nodes. Thus, GANI is unnoticeable enough and suitable for large ratio injections.

	\begin{table}[]
		\centering
		\renewcommand\arraystretch{1.1}
		\caption{Accuracy of GCN on each module with injecting 5\% nodes. Clean row is for reference. The best result is boldfaced.}
		\begin{tabular}{c | c c c c}
			\bottomrule
			\textbf{Modules}         & \textbf{Cora} & \textbf{Citeseer} & \textbf{Cora-ML} & \textbf{Pubmed}
			
			\\ \hline\hline
			Clean & 0.8360 & 0.7287 & 0.8536 & 0.8649\\ 
			Random & 0.8303 & 0.7250  & 0.8500 & 0.8648\\ 
			GANI$_{\rm -links}$ & 0.8300 & 0.7229   & 0.8494 & 0.8625\\ 
			GANI$_{\rm -features}$ & 0.7946 & 0.7066  & 0.8148 & \textbf{0.8356}\\
			GANI & \textbf{0.7725} & \textbf{0.7027}  & \textbf{0.8082} & 0.8392\\ 
			\bottomrule
		\end{tabular}
		
		\label{table:ablation}
	\end{table}

	\subsubsection{Ablation Study}
	Finally, to further verify the effectiveness of GANI, we analyze the influence of each part of GANI through ablation study. As our GANI majorly includes the feature generation module and neighbor selection module, we conduct an ablation study by removing one or all of them to verify their power on node injection attacks. The specific different combinations are as follows. Random indicates we will randomly decide the features and neighbors of the adversarial nodes. GANI$_{\rm -links}$ indicates we will retain the feature generation part of GANI, but randomly design the adversarial links on the neighbor selection procedure. GANI$_{\rm -features}$ represents that we will randomly assign the adversarial features on the feature generation part, but retain the GA-based neighboring selection procedure. 
	
	As shown in Table \ref{table:ablation}, compared with the feature generation part, the neighbor selection step contributes more to the final performance. The above results are reasonable since the adversarial link will decide the final aggregation target of nodes. If we cannot generate high-quality adversarial links, the malicious features will fail to contribute to attack performance. Also, the overall better performance of GANI indicates that the combination of appropriate features and neighbors can help improve the final performance. Particularly, we find that GANI with random features slightly outperforms the original GANI on Pubmed dataset. We believe the possible reason is that this setting (i.e., GANI$_{\rm -features}$) may randomly select the nodes whose non-zero indices have far exceeded our feature-attack budgets. Combining with our GA-based neighbor selection mechanism, this setting is possible to obtain comparable performance to GANI.

	\section{Conclusion}\label{sec:c}
	
	In this work, we study the global adversarial attacks on graph neural networks via node injections. Compared with previous studies, we design a more imperceptible yet effective node injection method, GANI, by considering the topology properties and the feature consistent aspects. Specifically, we first propose a degree sampling operation to retain the structural similarity of graphs. For the feature generation step, considering the difference between the binary and continuous features, we utilize the statistical information of nodes in each class to make the fake nodes similar to the original nodes in the feature domain. For the neighbor generation step, we select the optimal neighbors via the evolutionary perturbations obtained from genetic algorithm. Moreover, we induce a two-level sorting mechanism considering both the basic fitness function and the decrease of node homophily on the sorting procedure to further improve our method. Extensive experimental results demonstrate the superiority of the proposed GANI on the attack performance for different kinds of graph neural network models and the imperceptibility on both structure and feature properties. Particularly, the corresponding property investigation and outlier detection on the newly injected nodes also illustrate the imperceptible effect of the proposed method for the unnoticeable injections.


	\bibliographystyle{IEEEtran}
	\bibliography{IEEEabrv, mybib}

\end{document}